\documentclass[sigconf,nonacm]{acmart}
\AtBeginDocument{%
  }

\setcopyright{none}

\newcommand{\ind}{\perp\!\!\!\!\perp} 
\newcommand{\vecscalar}[1]{\boldsymbol{#1}}
\newcommand{\vecrv}[1]{\mathbf{#1}}
\newcommand{\rv}[1]{\mathrm{#1}}
\definecolor{AlgGreen}{RGB}{0,102,51} 
\definecolor{ICLRBlue}{HTML}{2F6FED}   
\definecolor{ICLRGreen}{HTML}{18A558}  
\newcommand{\zhu}[1]{\textcolor{ICLRBlue}{#1}}

\newtheorem{hypothesis}{Hypothesis}

\usepackage{ulem}
\usepackage{tcolorbox}
\renewenvironment{boxed}
    {
    \begin{tcolorbox}[colback=black!2!white,colframe=black!75!black,left=2pt,right=2pt,top=2pt,bottom=2pt]
    }
    { 
    \end{tcolorbox}
    }

\usepackage{multirow}
\usepackage{subfigure}
\usepackage{subcaption}
\usepackage{amsmath}
\usepackage{tikz}
\usepackage{enumitem}
\usepackage[table,xcdraw]{xcolor}
\usepackage{wrapfig}

\usepackage{algorithm}
\usepackage{algpseudocode}

\DeclareMathOperator*{\argmin}{arg\,min}

\begin{document}

\title{ELBO\textsubscript{TDS}: Temporal Distribution Shift Generalization for Industry-Scale Recommender Systems}

\makeatletter
\renewcommand{\@ACM@checkaffil}{%
  \if@ACM@instpresent\else
    \ClassWarningNoLine{\@classname}
      {No institution present for an affiliation}%
  \fi
}
\makeatother

\author{Yuxuan Zhu}
\affiliation{%
  \institution{Shopee}
}
\email{iamyuxuanzhu@gmail.com}

\author{Cong Fu}
\affiliation{%
  \institution{Shopee}
}
\email{fc731097343@gmail.com}

\author{Yabo Ni}
\affiliation{
 \institution{Nanyang Technological University} 
}
\email{niyabo1989@gmail.com}

\author{Anxiang Zeng}
\affiliation{
 \institution{Nanyang Technological University}
}
\email{zeng0118@e.ntu.edu.sg}

\author{Yuan Fang}
\affiliation{%
  \institution{Singapore Management University}
  }
\email{yfang@smu.edu.sg}
\authornote{Corresponding author.}


\begin{abstract}
Temporal distribution shift (TDS) degrades the generalization of recommender systems: models trained on past interactions face systematically different user behavior in future time windows. While incremental training on recent data is widely used in industry production, it can over-adapt to transient patterns and exacerbate instability. Existing approaches such as invariant learning and self-supervised learning offer partial solutions but often suffer from unstable temporal generalization, representation collapse, or inefficient data utilization. To address these limitations, we propose ELBO$_\text{TDS}$, a probabilistic framework for temporally robust recommendation. First, we identify key shifting factors on real-world production data and design a simple yet effective data augmentation strategy that simulate realistic time-varying factors to extend the training support. Second, we derive a self-supervised variational objective ELBO$_\text{TDS}$ to harness the benefits of this extended distribution while preventing representation collapse. Extensive experiments on both a public benchmark and a large-scale industrial dataset demonstrate ELBO$_\text{TDS}$ consistently improves next-period ranking performance over strong baselines, yielding a 2.33\% uplift in GMV per user online and has been successfully deployed in production. Code is available at \zhu{\url{https://github.com/FuCongResearchSquad/ELBO4TDS}}.
\end{abstract}


\maketitle

\begin{figure}[t]
  \includegraphics[width=0.85\linewidth]{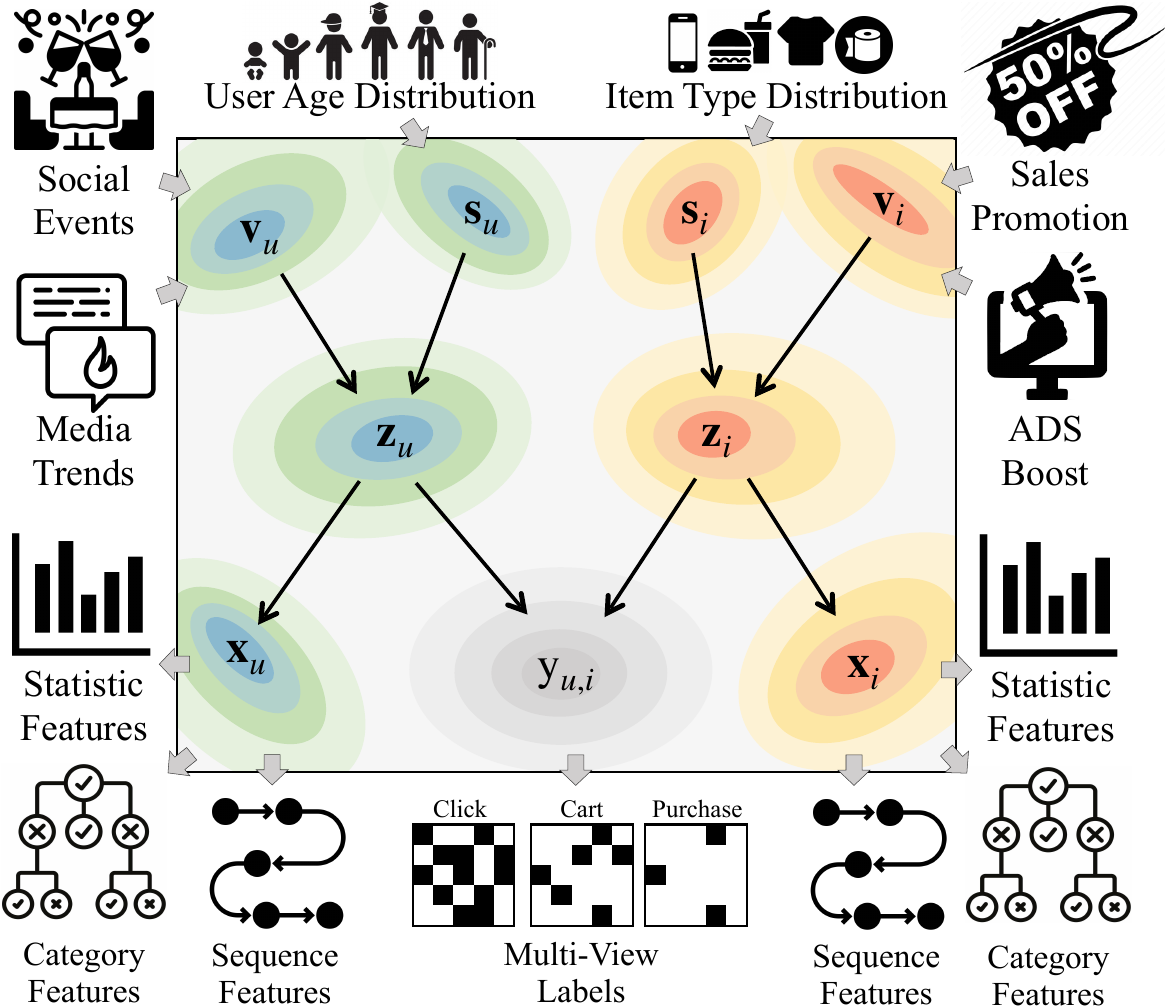}
  \caption{Data generating process to model temporal distribution shift. Subscript $u$ and $i$ indicate user- and item-specific variables, respectively. $\vecrv{v}$ denotes time-varying factors, $\vecrv{s}$ denotes relatively stable factors, $\vecrv{z}$ denotes latent variables, $\vecrv{x}$ denotes observed samples, and $\rv{y}$ denotes labels.} 
  \label{fig:causal-graph}
\end{figure}

\section{Introduction}
\label{sec:1}
Modern recommender systems learn from historical user--item interactions (e.g., clicks and purchases) to predict future engagement~\cite{ESMM,AITM,DCMT,Resflow}, implicitly
assuming historical and future interactions are independent and
identically distributed (I.I.D.).
However, user behavior evolves rapidly due to both internal interventions (e.g., flash sales, marketing campaigns) and external shocks (e.g., media influence, seasonality).
As a result, models trained on past data often face temporal distribution shift (TDS) at test time, leading to degraded next-period generalization.

A common industrial workaround is periodic incremental training, where models are fine-tuned on newly collected data (e.g., hourly or daily) rather than relying on a fixed checkpoint. While this approach helps capture emerging patterns, it also introduces instability. \textit{Models trained under rapidly shifting dynamics may fail to retain long-term preference signals, leading to noisy updates and brittle generalization.} Although TDS has attracted limited attention in prior work, two promising paradigms offer partial remedies:

\noindent \textbf{(1) Invariant learning}~\cite{arjovsky2020invariantriskminimization,pmlr-v139-krueger21a,yong2021empirical,Sagawa*2020Distributionally} aims to identify stable predictors across environments, yet defining environments along time is ambiguous (how to choose a proper time-window to separate enviroments in recommender systems) and many approaches require mixing or replaying historical data, which is expensive and can overfit to outdated patterns.

\noindent \textbf{(2) Self-supervised learning (SSL)}~\cite{zhang2018mixup,sui2023unleashing} with augmentation can expand training support, but common discriminative objectives (e.g., contrastive alignment) may suppress informative temporal variation and induce representation collapse~\cite{oord2019representationlearningcontrastivepredictive}, harming personalization—especially for sparse targets.

To formalize TDS, we model recommendation through the data generating process in Figure~\ref{fig:causal-graph}.
For each user-item pair, relatively stable factors $\vecrv{s}_u$ and $\vecrv{s}_i$ encode persistent user preferences and item properties, while time-varying factors $\vecrv{v}_u$ and $\vecrv{v}_i$ capture short-term effects such as promotions, trend shifts, and changing exposure. These variables jointly determine latent representations $\vecrv{z}_u$ and $\vecrv{z}_i$, which then generate observed features $\vecrv{x}_u$, $\vecrv{x}_i$, and the interaction label $\rv{y}_{u,i}$. 
Under this view, temporal distribution shift arises because the paths $\vecrv{v}_u\rightarrow\vecrv{z}_u\rightarrow\vecrv{x}_u$, $\vecrv{v}_i\rightarrow\vecrv{z}_i\rightarrow\vecrv{x}_i$, and $\vecrv{v}_u\rightarrow\vecrv{z}_u\rightarrow\rv{y}_{u,i}\leftarrow\vecrv{z}_i\leftarrow\vecrv{v}_i$ change over time, even when the stable factors remain relatively persistent within a short horizon. 
Our objective is to learn $\vecrv{z}$ preserve predictive information associated with $\vecrv{s}$ while reducing spurious dependence on $\vecrv{v}$. To achieve this, we inject controlled time-varying perturbations into training samples—guided by statistical insights from real production data, and formulate an Evidence Lower Bound (ELBO) objective that: \textbf{(i)} Filters out time-dependent noise by aligning the latent $\vecrv{z}$ of augmented views with the original; \textbf{(2)} Preserves both distinctive characteristics (of users and items) and predictive signals by jointly modeling $\vecrv{z} \to \vecrv{x}$ and $\vecrv{z} \to \rv{y}$.
Further, our augmentation strategy is designed with industry constraints in mind.
Based on empirical analyses of real-world feature drifts, we introduce lightweight, feature-aware perturbation (categorical, statistical, sequential) to simulate plausible temporal variations. These augmentations extend the training distribution and improve model robustness without additional online latency or storage cost.

ELBO$_\text{TDS}$ is modular and can be plugged into existing architectures, such as multitask learning \cite{MMoE,10.1145/3383313.3412236,10.1145/3447548.3467071,10.1145/3209978.3210104}, sequential models \cite{Granroth-Wilding_Clark_2016,10.1145/3269206.3271761,10.1145/3357384.3357895}, and list-wise ranking \cite{ListNet,10.1145/1390156.1390306}, with minor changes.

Our main contributions are summarized as follows:
\noindent \textbf{(1) A Unified Probabilistic Framework for TDS}: We present ELBO$_\text{TDS}$, combining feature-aware temporal augmentation with a reconstruction-regularized variational objective to encourage invariance to transient temporal factors while preserving predictive, sample-specific information.
\noindent \textbf{(2) Robust Industrial Integration:} We introduce a simple yet effective augmentation strategy tailored to real-world feature dynamics. Our method improves GMV/user by 2.33\% in online A/B tests and is fully deployed in a real-world search platform.

\section{Problem Definition}
\label{sec:2}
We consider the challenge of recommendation under TDS, where both user and item variables evolve over time due to various factors. 
At any time span $t$ (e.g., a natural day or hour), we observe a dataset $\mathcal{D}_t=\{(\vecrv{x}_{u,i,t},\rv{y}_{u,i,t})\}$, where $\vecrv{x}_{u,i,t}$ denotes an observed sample (a feature set) for user $u$ and item $i$ at $t$, and $\rv{y}_{u,i,t}$ is the interaction label (e.g., click, purchase) associated with ($u, i$) during $t$.

The temporal nature of the problem imposes the following constraints. \textbf{(1) Periodic Training}: The model is trained periodically on a sequence of datasets $\{\mathcal{D}_t\}_{t\le T}$, which are collected up to a current time span $T$. \textbf{(2) Future Prediction}: After training, the model is deployed to predict user behavior for the next time span $T+1$. Importantly, data from $T+1$ remains \textbf{unavailable for training} and can only be accessed once the time window $T+1$ ends.

\begin{figure*}[h]
    \centering
    \includegraphics[width=1.0\linewidth]{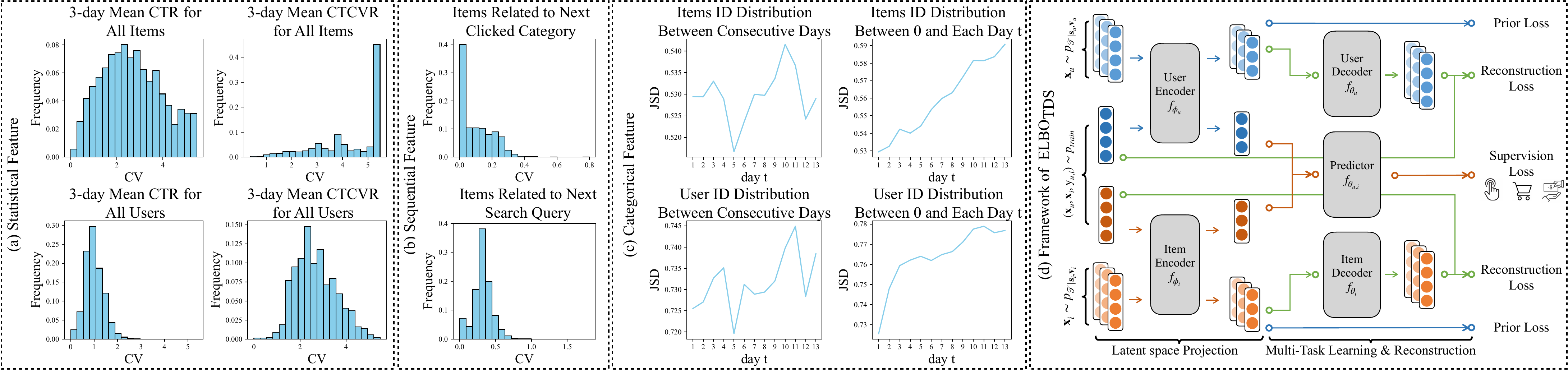}
    \caption{The statistical analysis of the distribution shift for statistical (a), sequential (b), and categorical (c) features, showing significant fluctuations and contributing to the sample TDS. Right side (d) illustrates the architecture of ELBO$_\text{TDS}$.}
    \label{fig:dist_shift_arch}
\end{figure*}

\section{ELBO$_{\text{TDS}}$: TDS Generalization}
In this section, we present ELBO$_{\text{TDS}}$ for temporal distribution shift generalization. We first summarize temporal drift in production data with a stable-drifting hypothesis. We then model TDS through the data generating process in Figure 1 and derive a variational objective that, together with feature-aware augmentation, learns representations invariant to transient temporal factors while preserving predictive information.

\subsection{Empirical Observation of TDS}
\label{sec:3.1}

Before introducing ELBO$_{\text{TDS}}$, we first examine how production data drift over time in a large-scale recommender. We focus on three common types: statistical, sequential, and categorical features.

Our analysis reveals that different feature types exhibit distinct patterns of temporal variation. To systematically quantify these shifts, we adopt two key metrics: the Coefficient of Variation (CV), which measures the relative variability of a feature with respect to its mean, and the Jensen–Shannon Divergence (JSD), which quantifies the distributional divergence between two temporal snapshots. The results, visualized in Figure~\ref{fig:dist_shift_arch}, yield three core insights: \textbf{(1)} All features fluctuate over time, but the degree of variation differs significantly; \textbf{(2)} For most items and users, these fluctuations remain bounded. For example, the 3-day mean click-through rate (CTR) of most items exhibits a CV less than 4, suggesting "stable" short-term dynamics. \textbf{(3)} While short-term variations are prominent, they do not accumulate significantly over time. This is evidenced by JSD values in Figure~\ref{fig:dist_shift_arch}(c), where the divergence between day 0 and day 13 is only marginally higher than that between day 0 and day 1.

These empirical observations motivate the following hypothesis on TDS in recommender, which underpins our method design:
\begin{boxed}
\begin{hypothesis}[Stable-Drifting Distribution Hypothesis]
Within a manageable time window, user preferences and item characteristics are governed by stable latent factors $\vecrv{s}$, while separable time-varying factors $\vecrv{v}$ induce short-term fluctuations around a stable distributional core.
\label{hypo:tds}
\end{hypothesis}
\end{boxed}

\subsection{Derivation of ELBO$_{\text{TDS}}$}
\label{sec:3.2}

Building on Hypothesis \ref{hypo:tds}, we revisit the data generation process and introduce a corresponding optimization framework.

First, we present a unified data-generation process (Figure~\ref{fig:causal-graph}) that captures the causal dependencies in recommender systems under temporal distribution shift (TDS) settings.
\begin{itemize}[leftmargin=*]

    \item $\vecrv{s}_u$ and $\vecrv{s}_i$: the abstract aggregated variables that contains the semantically stable information of user and item, respectively, e.g., the category of an item, which is relatively stable over time.
    \item $\vecrv{v}_u$ and $\vecrv{v}_i$: the abstract aggregated time-varying variable for user $u$ and item $i$, respectively, e.g., fluctuated item sales counts.
    \item $\vecrv{z}_u$ and $\vecrv{z}_i$: the latent variable (a.k.a., representation) for user $u$ and item $i$, respectively, determined by both $\vecrv{s}$ and $\vecrv{v}$.
    \item $\vecrv{x}_u$ and $\vecrv{x}_i$: the observed feature (covariate) of user $u$ and item $i$, respectively, determined by the latent variable $\vecrv{z}_u$ and $\vecrv{z}_i$.
    \item $\rv{y}_{u,i}$: the interaction variable (e.g. click or conversion) between user $u$ and item $i$, determined by both $\vecrv{z}_u$ and $\vecrv{z}_i$.

\end{itemize}
From this process, we see three main paths driving TDS: 1) $\vecrv{v}_u\rightarrow\vecrv{z}_u\rightarrow\vecrv{x}_u$ leads to the shift in user covariates $\vecrv{x}_u$; 2) $\vecrv{v}_i\rightarrow\vecrv{z}_i\rightarrow\vecrv{x}_i$ leads to the shift in item covariates $\vecrv{x}_i$; 3) $\vecrv{v}_u\rightarrow\vecrv{z}_u\rightarrow\rv{y}_{u,i}\leftarrow\vecrv{z}_i\leftarrow\vecrv{v}_i$ can lead to spurious correlations captured by a model. Collectively, these paths induce the joint distribution shift: $p_{T-1}(\vecrv{x}_u,\vecrv{x}_i,\rv{y}_{u,i})\neq p_{T}(\vecrv{x}_u,\vecrv{x}_i,\rv{y}_{u,i})$, violating the i.i.d. assumption fundamental to Empirical Risk Minimization (ERM). For example, consider the scenario in Figure~\ref{fig:causal-graph}, on a regular day ($T-1$), users rarely purchase overpriced luxury items. Next, on a promotion day ($T$), heavy discounts and advertising ($\vecrv{v}$) significantly increase their sales. ERM models trained on $p_{T-1}$ memorizes past patterns and fails to adapt, mispredicting user behaviors at promotion when pricing dynamics shift.

To improve generalization under temporal distribution shift (TDS), our objective is to retain the stable, informative factors while filtering out time-varying noise. This is enabled by an extended training distribution, constructed via augmentation to simulate realistic shifting patterns. Formally, grounded in the causal graph presented in Figure~\ref{fig:causal-graph}, we define the extended distribution as: $p_{\mathcal{T}|\vecrv{s}_u,\vecrv{s}_i,\vecrv{v}_u,\vecrv{v}_i}$, where $\mathcal{T}$ denotes the augmented temporal support designed to reflect distributional changes (details provided in Section~\ref{sec:3.3}). Under this data generating process, our learning objective becomes maximizing the conditional likelihood given the stable factors $\vecrv{s}$ under the enriched distribution. 
\begin{equation}
\label{eq:likelihood}
\mathop{\max}_{\theta}{\mathbb{E}_{(\vecrv{x}_u,\vecrv{x}_i,\rv{y}_{u,i},\vecrv{s}_u,\vecrv{s}_i)\sim p_{\mathcal{T}|\vecrv{s}_u,\vecrv{s}_i,\vecrv{v}_u,\vecrv{v}_i}}[\mathop{\log}p_{\theta}(\vecrv{x}_u,\vecrv{x}_i,\rv{y}_{u,i}|\vecrv{s}_u,\vecrv{s}_i)]},
\end{equation}
where $\vecrv{s}_u$ and $\vecrv{s}_i$ denote the stable latent factors governing the generation of user and item covariates, respectively. These factors determine which samples share the same underlying semantics, and are implicitly realized through the augmentation strategy $\mathcal{T}$. Similar to self-supervised learning, we treat augmented views of the same base sample as sharing the same stable factors $\vecrv{s}$, whereas views of different base samples within a batch are assumed to correspond to different $\vecrv{s}$. This design encourages the model to preserve invariant semantic content while filtering out sample-specific temporal noise introduced through $\vecrv{v}_u$ and $\vecrv{v}_i$.

To optimize Eqn.~\eqref{eq:likelihood}, we derive an Evidence Lower Bound \cite{bizeul2024a} based on the data generating process in Figure~\ref{fig:causal-graph} (see derivation in Appendix~\ref{app:proof-elbo}) as:
\begin{equation}
\label{eq:elbo}
\begin{aligned}
&\mathop{\max}_{\theta}{\mathbb{E}_{(\vecrv{x}_u,\vecrv{x}_i,\rv{y}_{u,i},\vecrv{s}_u,\vecrv{s}_i)\sim p_{\mathcal{T}|\vecrv{s}_u,\vecrv{s}_i,\vecrv{v}_u,\vecrv{v}_i}}[\mathop{\log}p_{\theta}(\vecrv{x}_u,\vecrv{x}_i,\rv{y}_{u,i}|\vecrv{s}_u,\vecrv{s}_i)]} \\
\geq & \mathop{\max}_{\theta, \phi}\mathbb{E}_{(\vecrv{x}_u,\vecrv{x}_i,\rv{y}_{u,i},\vecrv{s}_u,\vecrv{s}_i)\sim p_{\mathcal{T}|\vecrv{s}_u,\vecrv{s}_i,\vecrv{v}_u,\vecrv{v}_i}}[\mathbb{E}_{\vecrv{z}_u\sim q_{\phi}(\vecrv{z}_u|\vecrv{x}_u),\vecrv{z}_i\sim q_{\phi}(\vecrv{z}_i|\vecrv{x}_i)}[ \\
&\underbrace{\mathop{\log} p_{\theta}(\vecrv{x}_u|\vecrv{z}_u) + \mathop{\log}p_{\theta}(\vecrv{x}_i|\vecrv{z}_i)}_{\text{reconstruction term}} \underbrace{-\mathop{\log}q_{\phi}(\vecrv{z}_u|\vecrv{x}_u)-\mathop{\log}q_{\phi}(\vecrv{z}_i|\vecrv{x}_i)}_{\text{entropy term}}\\
& + \underbrace{\mathop{\log} p_{\theta}(\rv{y}_{u,i}|\vecrv{z}_u,\vecrv{z}_i)}_{\text{predictive term}} + \underbrace{\mathop{\log} p(\vecrv{z}_u|\vecrv{s}_u)+ \mathop{\log} p(\vecrv{z}_i|\vecrv{s}_i)}_{\text{prior term}}]]
\end{aligned}
\end{equation}
where $q_{\phi}(\vecrv{z}_u|\vecrv{x}_u)$ and $q_{\phi}(\vecrv{z}_i|\vecrv{x}_i)$ are the approximated posteriors of $p(\vecrv{z}_u|\vecrv{x}_u)$ and $p(\vecrv{z}_i|\vecrv{x}_i)$ parameterized by $\phi$.
The ELBO can be interpreted as four terms: \textit{reconstruction, entropy, predictive, and prior terms}. 
For clarity, we group the reconstruction term, entropy term, and prior term into a \textbf{self-supervised component}, while the predictive term constitutes the \textbf{supervised component}.

\subsubsection{Self-Supervised Component} We now provide explanations for each term and introduce tractable optimization objectives to construct the loss function for the self-supervised component.

\noindent\textbf{Prior term.} To ensure robustness against TDS, we enforce view-invariant representation learning. For a given user-item pair $(u,i)$, various data views arise due to different strengths of time-varying factors $\vecrv{v}_u,\vecrv{v}_i$. The prior term ensures that representations from all such views collapse toward the same prior distributions: $p(\vecrv{z}_u|\vecrv{s}),p(\vecrv{z}_i|\vecrv{s})$. This process eliminates view-specific temporal noise, encouraging $\vecrv{z}_u$ and $\vecrv{z}_i$ capture only time-invariant user and item characteristics.

Formally, we assume $q_{\phi}(\vecrv{z}_u|\vecrv{x}_u)=\mathcal{N}(f_{\phi_{u}}(\vecrv{x}_u),\mathbf{I})$ and $q_{\phi}(\vecrv{z}_i|\vecrv{x}_i)=\mathcal{N}(f_{\phi_{i}}(\vecrv{x}_i),\mathbf{I})$. Here, $f_{\phi_{u}}$ and $f_{\phi_{i}}$ are neural networks that estimate the mean representations, while the variance is fixed to $\mathbf{I}$ for training stability. For efficient sampling, we apply the reparameterization trick \cite{DBLP:journals/corr/KingmaW13} with auxiliary noise $\epsilon_u\sim\mathcal{N}(\mathbf{0},\mathbf{I})$ and $\epsilon_i\sim\mathcal{N}(\mathbf{0},\mathbf{I})$, yielding: $\vecrv{z}_u=f_{\phi_{u}}(\vecrv{x}_u)+\mathbf{I}\epsilon_u$ and $\vecrv{z}_i=f_{\phi_{i}}(\vecrv{x}_i)+\mathbf{I}\epsilon_i$. Further, we assume the prior distributions $p(\vecrv{z}_u|\vecrv{s}_u)$ and $p(\vecrv{z}_i|\vecrv{s}_i)$ follow Gaussian distributions. Under this assumption, the prior loss is given by:
\begin{equation}
\label{eq:prior_loss}
    \mathcal{L}_{prior} = \mathcal{L}_{MSE}(\vecrv{z}_u, \vecrv{\overline{z}}_u) + \mathcal{L}_{MSE}(\vecrv{z}_i, \vecrv{\overline{z}}_i),
\end{equation}
where $\vecrv{\overline{z}}_u = \frac{1}{J}\sum_{j=1}^{J}\vecrv{z}_{u}^{j}$ and $\vecrv{\overline{z}}_i = \frac{1}{J}\sum_{j=1}^{J}\vecrv{z}_{i}^{j}$, averaged over $J$ data views $\{\vecrv{x}_u^j,\vecrv{x}_i^j\}_{j=1}^J$. The derivation of Eqn.~\eqref{eq:prior_loss} is given in Appendix~\ref{app:proof_prior}.

\noindent\textbf{Entropy term} 
controls the variance the the latent variable $\vecrv{z}$.
Keeping a proper entropy can help avoid the collapse of $\vecrv{z}$ to some extent \cite{bizeul2024a}.
Since $\vecrv{z}_u$ and $\vecrv{z}_i$ are Gaussian with unit variance (as mentioned in prior term), the entropy becomes a constant and contributes 0 to the gradient, thus can be canceled out.

\noindent\textbf{Reconstruction term} ensures that the latent representations $\vecrv{z}_u$ and $\vecrv{z}_i$ retain individual-specific information by reconstructing the original user and item features $\vecrv{x}_u$ and $\vecrv{x}_i$. To achieve this, we employ deterministic neural networks $f_{\theta_{u}}$ and $f_{\theta_{i}}$ to to reconstruct the input covariates from the latent space: $\hat{\vecrv{x}}_u=f_{\theta_{u}}(\vecrv{z}_u)$ and $\hat{\vecrv{x}}_i=f_{\theta_{i}}(\vecrv{z}_i)$. We optimize the reconstruction objective using the MSE loss:
    \begin{equation}
        \mathcal{L}_{recon} = \mathcal{L}_{MSE}(\hat{\vecrv{x}}_u, \vecrv{x}_u) + \mathcal{L}_{MSE}(\hat{\vecrv{x}}_i, \vecrv{x}_i).
    \end{equation}

Combining them yields the self-supervised objective:
\begin{equation}
    \mathop{\min}_{\phi, \theta}\mathbb{E}_{(\vecrv{x}_u,\vecrv{x}_i,\vecrv{s}_u,\vecrv{s}_i)\sim p_{\mathcal{T}|\vecrv{s}_u,\vecrv{s}_i,\vecrv{v}_u,\vecrv{v}_i}}[\mathcal{L}_{recon} + \mathcal{L}_{prior}].
\end{equation}

\subsubsection{Supervised Component} The \textbf{predictive term} aligns latent representations $\vecrv{z}_u$ and $\vecrv{z}_i$ with the target interaction $\rv{y}_{u,i}$, which can be interpreted as any recommendation tasks (e.g., CTR prediction) modeled by corresponding deterministic network $f_{\theta_{u,i}}$. To ensure the training-inference consistency and ranking capability, we use the deterministic means $f_{\phi_{u}}(\vecrv{x}_u)$ and $f_{\phi_{i}}(\vecrv{x}_i)$ (extracted from truly observed data $\vecrv{x}_u,\vecrv{x}_i$). The supervised objective then is given by :
\begin{equation}
    \mathop{\min}_{\phi, \theta}\mathbb{E}_{(\vecrv{x}_u,\vecrv{x}_i,\rv{y}_{u,i})\sim p_{train}}[\mathcal{L}_{pred}],
\end{equation}
where $\mathcal{L}_{pred}=\mathcal{L}_{pred}(f_{\theta_{u,i}}(f_{\phi_{u}}(\vecrv{x}_u),f_{\phi_{i}}(\vecrv{x}_i)),\rv{y}_{u,i})$ is the task-specific loss, and $p_{train}$ is the truly observed distribution (subset of $\mathcal{T}$).

Combining the self-supervised and supervised components  yields the total loss of our framework:
\begin{equation}
\label{eq:overall_loss}
\begin{aligned}
&\mathop{\min}_{\phi, \theta}\mathbb{E}_{(\vecrv{x}_u,\vecrv{x}_i,\rv{y}_{u,i})\sim p_{train}}[\mathcal{L}_{pred}] \\
&+ \alpha\mathbb{E}_{(\vecrv{x}_u,\vecrv{x}_i,\vecrv{s}_u,\vecrv{s}_i)\sim p_{\mathcal{T}|\vecrv{s}_u,\vecrv{s}_i,\vecrv{v}_u,\vecrv{v}_i}}[\mathcal{L}_{recon} + \mathcal{L}_{prior}],
\end{aligned}
\end{equation}
where $\phi=(\phi_{u},\phi_{i})$, $\theta=(\theta_{u},\theta_{i},\theta_{u,i})$, and $\alpha$ is a hyperparameter that balances the self-supervised and supervised components.

\subsection{Data Augmentation and Training Pipeline}
\label{sec:3.3}
Under the data generating process in Figure 1, temporal distribution shift is induced by changes in the time-varying factors $\vecrv{v}$ while the stable factors $\vecrv{s}$ remain relatively persistent within a short horizon.
Therefore, we design a data augmentation strategy that generates $J$ augmented views $(\vecrv{x}_{u}^{j},\vecrv{x}_{i}^{j})_{j=1}^{J} = \mathcal{T}(\vecrv{x}_{u},\vecrv{x}_{i}|\vecrv{s}_u,\vecrv{s}_i,\vecrv{v}_u,\vecrv{v}_i)$,
with two goals: (1) approximate realistic temporal fluctuations and (2) preserve a relatively stable core.
Guided by the empirical findings in Section~\ref{sec:3.1}, 
we design feature-aware augmentations for three common feature types.

\noindent\textbf{Statistical Features.} fluctuate around a mean. To mimic this, we perturb them within a controlled range while keeping their global distribution intact. Because different statistics have different scales, we first bucketize each feature into equal-width bins. Given an original bucket index $B_i$, we sample a perturbed index:
$
B_i' = \min(\max(B_i+\epsilon, 0), B_{max}) \quad \text{s.t.} \quad \epsilon\sim \mathcal{U}\{-r,r\},
$
where $B_{\max}$ is the max bin index and $r$ controls perturbation strength.

\noindent\textbf{Sequential Features.} Historical action sequences also fluctuate in length and content. To reflect this, we adopt random masking (akin to masked language modeling \cite{devlin-etal-2019-bert,liu2020roberta}) so the model learns contextual signals rather than memorizing specific subsequences. Concretely, we sample a binary mask $\vecrv{m} \sim \operatorname{Bern}(p_{seq})$, where $p_{seq}$ is the masking probability.

\noindent\textbf{Categorical Features.} Categorical distributions drift continuously over time, risking overfitting to transient modes. We therefore apply embedding dropout with rate $p_{cate}$, randomly zeroing embeddings to prevent reliance on any single categorical snapshot.

\noindent\textbf{Global Perturbation Selection.} 
To avoid overly aggressive augmentation that would push samples outside realistic temporal support, we perturb only a random subset of features per sample. This “global perturbation cap,” inspired by vision augmentations such as Random Erasing and Cutout \cite{10.5555/3327345.3327439,NIPS2016_371bce7d}, keeps augmented data faithful to real-world TDS patterns.

\noindent\textbf{Training Pipeline.} Our augmentations are lightweight, we generate $J$ augmented views on the fly for each batch. 
The resulting ELBO$_\text{TDS}$ can plug seamlessly into incremental training pipelines with minimal code changes (see Figure~\ref{fig:dist_shift_arch}(d) and Algorithm~\ref{alg:elbo}).

\subsection{Collapse Analysis of InfoNCE}
\label{sec:3.4}
In this section, we theoretically compare ELBO$_{\text{TDS}}$
with InfoNCE, offering intuition for why ELBO$_{\text{TDS}}$ can perform better under TDS. We present a lower bound for InfoNCE, stated in Proposition~\ref{prop:infonce_upper_bound}.
\begin{boxed}
\begin{proposition}
\label{prop:infonce_upper_bound}
Let $\vecrv{x}_u$ and $\vecrv{x}_i$ be user and item features with their respective augmented views $\vecrv{X}_u=\{\vecrv{x}_u,\vecrv{x}^{+,0}_u,\vecrv{x}^{-,1}_u,..,\vecrv{x}^{-,k}_u\}$ and $\vecrv{X}_i=\{\vecrv{x}_i,\vecrv{x}^{+,0}_i,\vecrv{x}^{-,1}_i,..,\vecrv{x}^{-,k}_i\}$. the InfoNCE loss upper bounds the following expression:
\begin{equation*}
\begin{aligned}
\mathcal{L}_{\text{INCE}}^{u} + \mathcal{L}_{\text{INCE}}^{i}\geq& -\mathbb{E}_{\vecrv{x}_u, \vecrv{x}_i}[\mathbb{E}_{\vecrv{z}_u, \vecrv{z}_i}[\log p(\vecrv{z}_u|\vecrv{s}_u) + \log p(\vecrv{z}_i|\vecrv{s}_i) \\
&- \log p(\vecrv{z}_u) - \log p(\vecrv{z}_i)]] + c
\end{aligned}
\end{equation*}
where $c=2\log k$ and k is the number of negative samples.
\end{proposition}
\end{boxed}
\noindent The proof is given in Appendix~\ref{sec:proof-prop1}.
Proposition~\ref{prop:infonce_upper_bound} shows that InfoNCE actually minimizes an upper bound of an ELBO‑like objective, where the reconstruction term is replaced by entropy penalties $-\log p(\vecrv{z}_u)$ and $-\log p(\vecrv{z}_i)$. Although these entropy terms promote uniformity and discourage severe over-clustering, they do not explicitly enforce the retention of per-sample, idiosyncratic details, leaving room for harmful representation collapse.

In e-commerce recommendation, such fine-grained details often prove crucial for downstream tasks, such as personalized ranking. As an example, suppose multiple users share a preference for iPhones (a common semantic theme), yet some users are notably price-insensitive, preferring new models over older versions. If the representation learner only clusters iPhone fans as a group without constraints on personalized information preservation, it may overlook these individual nuances—leading to less accurate recommendations. By contrast, a reconstruction-based approach retains these subtle distinctions, enabling more precise personalization.

\noindent\textbf{Remark.}
In contrast to InfoNCE, the reconstruction terms $p_{\theta}(\vecrv{x}_u|\vecrv{z}_u)$ and $p_{\theta}(\vecrv{x}_i|\vecrv{z}_i)$ of ELBO$_\text{TDS}$ promote individual-level feature preservation. This preservation is key to precise item recommendations.

\section{Experiment}
In this section, we conduct extensive experiments in both offline and online settings.

\begin{table}[t]
    \centering
    \caption{Dataset statistics.}
    \vspace{-2mm}
    \setlength{\tabcolsep}{2.0pt}
    \resizebox{\linewidth}{!}{
\begin{tabular}{lccccc}
\hline
Dataset & \#Users & \#Items & Feat (U/I) & Labels (pos.\%) & Split Sizes \\ \hline
KuaiRand-1K &
1K &
2.4M &
\begin{tabular}[c]{@{}c@{}}30 / 63 total\\ stat: 0 / 52\\ cate: 30 / 11\end{tabular} &
\begin{tabular}[c]{@{}c@{}}\texttt{is\_like} / \texttt{is\_follow}\\ 1.9 / 0.1\end{tabular} &
\begin{tabular}[c]{@{}c@{}}train 5.3M\\ val 0.7M\\ test 0.7M\end{tabular} \\ \hline
Commercial-Small &
6.1M &
56M &
\begin{tabular}[c]{@{}c@{}}101 / 105 total\\ stat: 59 / 93\\ seq: 35 / 1\\ cate: 7 / 11\end{tabular} &
\begin{tabular}[c]{@{}c@{}}click / atc / purchase\\ 14.4 / 1.4 / 0.4\end{tabular} &
\begin{tabular}[c]{@{}c@{}}train 0.7B\\ val 60M\\ test 60M\end{tabular} \\ \hline
Commercial-Main &
-- &
-- &
\begin{tabular}[c]{@{}c@{}}101 / 105 total\\ stat: 59 / 93\\ seq: 35 / 1\\ cate: 7 / 11\end{tabular} &
\begin{tabular}[c]{@{}c@{}}click / atc / purchase\\ 11 / 1 / 0.2\end{tabular} &
\begin{tabular}[c]{@{}c@{}}train 28B\\ val 0.9B\\ test 0.9B\end{tabular} \\ \hline
\end{tabular}
}
    \label{tab:stat_datasets}
    \vspace{-2mm}
\end{table}

\subsection{Experiment Settings}

\subsubsection{Datasets}
We evaluate under a chronological protocol on one public benchmark and one large-scale proprietary dataset. The dataset statistics are given in Table~\ref{tab:stat_datasets}.

\noindent\textbf{(i) Commercial.} Interaction logs from a large-scale e-commerce search recommender system, spanning 40 consecutive days with 206 engineered features (statistical, sequential, categorical) and three feedback signals (click, add-to-cart, purchase). We also use a smaller subset named \textbf{Commercial-Small}\footnote{\zhu{\url{https://huggingface.co/datasets/PIIR/TDSRec-dataset}}} consisting of 13 consecutive days to enable reproducible experiments.

\noindent\textbf{(ii) KuaiRand-1K\footnote{\zhu{\url{https://kuairand.com/}}}.} A public dataset with rich feature semantics and timestamps, suitable for studying temporal shift.

\noindent\textbf{Splits.} For the commercial dataset (and its subset), we use the last day for test, the second-last day for validation, and the remaining days for training. For KuaiRand-1K, we split data into 10 equal-sized time spans, using the last span for test, the second-last for validation, and the rest for training.

\subsubsection{Baselines}
We compare ELBO$_{\text{TDS}}$ with strong baseline methods:
\textbf{ERM} (incremental empirical risk minimization), \textbf{Aug} (training directly on augmented data w/o ELBO$_{\text{TDS}}$),
invariant-learning baselines \textbf{IRM}~\cite{arjovsky2020invariantriskminimization}, \textbf{V-REx}~\cite{pmlr-v139-krueger21a}, \textbf{RVP}~\cite{yong2021empirical},
and self-supervised baselines \textbf{SSL4Rec}~\cite{10.1145/3459637.3481952}, \textbf{SSL4Rec*} (full-feature augmentation variant), and \textbf{Dino4Rec} (auxiliary self-distillation task inspired by DINOv2~\cite{oquab2023dinov2}).
 
\begin{table}[]
    \centering
    \caption{Multi-Task listwise ranking on Commercial-Small on MLP backbone and Cross-entropy\&ListNet loss.}
    \vspace{-2mm}
    \resizebox{\linewidth}{!}{
\begin{tabular}{lcccccc}
\hline
\multirow{2}{*}{\begin{tabular}[c]{@{}l@{}}backbone: MLP\\ loss: CE\&ListNet \cite{ListNet} \end{tabular}} & \multicolumn{2}{c}{Click}           & \multicolumn{2}{c}{Add To Cart}           & \multicolumn{2}{c}{Purchase}         \\
                                                                                                     & AUC             & GAUC            & AUC             & GAUC            & AUC             & GAUC            \\ \hline
ERM                                                                                                  & 0.7977          & 0.7910          & 0.8494          & 0.7966          & 0.8920          & 0.8228          \\
Aug                                                                                                  & 0.7982          & 0.7905          & 0.8497          & 0.7979          & 0.8917          & 0.8249          \\
IRM                                                                                                  & 0.7775          & 0.7715          & 0.8094          & 0.7624          & 0.8421          & 0.7831          \\
V-REx                                                                                                & 0.7827          & 0.7764          & 0.8155          & 0.7708          & 0.8446          & 0.7898          \\
RVP                                                                                                  & 0.7821          & 0.7756          & 0.8134          & 0.7676          & 0.8419          & 0.7838          \\
DINO4Rec                                                                                             & 0.7996          & 0.7914          & 0.8475          & 0.7961          & 0.8904          & 0.8218          \\
SSL4Rec                                                                                     & 0.8005          & 0.7932          & 0.8496          & 0.7973          & 0.8906          & 0.8213          \\
SSL4Rec*                                                                                              & 0.8012          & 0.7941          & 0.8490          & 0.7985          & 0.8894          & 0.8228          \\
ELBO$_{\text{TDS}}$                                                                                  & \textbf{0.8042} & \textbf{0.7976} & \textbf{0.8528} & \textbf{0.8037} & \textbf{0.8934} & \textbf{0.8297} \\ \hline 
ELBO$_{\text{TDS}}$+stat                                                                             & 0.8040          & 0.7974          & 0.8511          & 0.8021          & 0.8935          & 0.8294          \\
ELBO$_{\text{TDS}}$+seq                                                                              & 0.8036          & 0.7967          & 0.8520          & 0.8029          & 0.8930          & 0.8295          \\
ELBO$_{\text{TDS}}$+cate                                                                             & 0.8020          & 0.7957          & 0.8510          & 0.8019          & 0.8935          & 0.8285          \\ \hline
\end{tabular}
}
    
    \label{tab:auc_shopee}
\vspace{-2mm}
\end{table}

\subsubsection{Evaluation Metrics}
We adopt AUC and GAUC for pointwise~\cite{ESMM,BPR} and listwise~\cite{ListNet} ranking.
We adopt GMV/User (Gross Merchandise Value) to assess business impact in online A/B tests.

\subsubsection{Implementations}
All methods share the same backbone (twin-tower architecture~\cite{tan2023optimizing}) for fair comparison. We tune hyperparameters via grid search and additional details are given in Appendix~\ref{app:hyper-param}.

\begin{table}[]
    \centering
    \caption{Single- and multi-task ranking performance on Kuairand-1K on MLP backbone and cross entropy loss.}
    \vspace{-2mm}
    \setlength{\tabcolsep}{1.0pt}
\resizebox{1.0\linewidth}{!}{
\begin{tabular}{lcccccc}

\hline
\multirow{3}{*}{\begin{tabular}[c]{@{}l@{}}MLP\\ Loss: CE\end{tabular}} & \multicolumn{2}{c}{KuaiRand-1K single-task}     & \multicolumn{4}{c}{KuaiRand-1K multi-task}                                                        \\
                                                                                   & \multicolumn{2}{c}{is\_like}                    & \multicolumn{2}{c}{is\_like}                    & \multicolumn{2}{c}{is\_follow}                  \\
                                                                                   & AUC                    & GAUC                   & AUC                    & GAUC                   & AUC                    & GAUC                   \\ \hline
ERM                                                                                & 0.9053±0.0029          & 0.5086±0.0053          & 0.9015±0.0029          & 0.5025±0.0042          & 0.7816±0.0134          & 0.5016±0.0014          \\
Aug                                                                                & 0.9032±0.0035          & 0.5013±0.0044          & 0.8947±0.0032          & 0.5009±0.0046          & 0.7952±0.0045          & 0.4997±0.0039          \\
IRM                                                                                & 0.9043±0.0019          & 0.5079±0.0027          & 0.9034±0.0008          & 0.5008±0.0044          & 0.7816±0.0161          & 0.5016±0.0021          \\
V-REx                                                                              & 0.9046±0.0013          & 0.5060±0.0034          & 0.9034±0.0004          & 0.5022±0.0034          & 0.7779±0.0204          & 0.5008±0.0003          \\
RVP                                                                                & 0.9049±0.0022          & 0.5093±0.0047          & 0.9035±0.0040          & 0.5020±0.0033          & 0.7783±0.0113          & 0.5011±0.0015          \\
Dino4Rec                                                                            & 0.9046±0.0017          & 0.5066±0.0034          & 0.9017±0.0017          & 0.5020±0.0041          & 0.7841±0.0032          & 0.5011±0.0013          \\
SSL4Rec                                                                            & 0.9001±0.0007          & 0.5042±0.0066          & 0.9014±0.0024          & 0.5015±0.0062          & 0.7837±0.0061          & 0.5020±0.0024          \\
SSL4Rec*                                                                            & 0.9072±0.0006          & 0.5020±0.0054          & 0.9060±0.0025          & 0.4986±0.0042          & 0.8101±0.0050          & 0.4998±0.0047          \\
ELBO$_{\text{TDS}}$                                                                & \textbf{0.9112±0.0008} & \textbf{0.5191±0.0037} & \textbf{0.9093±0.0011} & \textbf{0.5101±0.0065} & \textbf{0.8260±0.0030} & \textbf{0.5024±0.0013} \\ \hline
\end{tabular}}
    
    \label{tab:auc_kuairand}
\end{table}

\subsection{Offline Results and Analyses}

\subsubsection{\textbf{RQ1\&2: Overall Performance \& Downstream Task Evaluation}}
We evaluate ELBO$_{\text{TDS}}$ on three settings: single-task and multi-task pointwise ranking on KuaiRand-1K, and multi-task listwise ranking on Commercial-Small.
Tables~\ref{tab:auc_kuairand} and~\ref{tab:auc_shopee} show:

\begin{itemize}[leftmargin=*]
\item \textbf{Consistent gains under TDS.} ELBO$_{\text{TDS}}$ achieves the best AUC and GAUC across datasets and tasks, indicating strong next-period generalization.
\item \textbf{Invariant learning is unstable under incremental updates.} IRM/V-REx/RVP provide limited gains on KuaiRand-1K and can underperform ERM on Commercial-Small, consistent with the difficulty of identifying truly time-invariant factors and the tendency to overfit when revisiting past windows~\cite{zhang2022towards,hsu2024taming}.
\item \textbf{SSL helps dense targets but can hurt sparse ones.} SSL baselines improve dense feedback (e.g., click, \texttt{is\_like}) but show weaker or negative gains on sparse targets (e.g., purchase). This is consistent with representation collapse suppressing rare but informative signals (Section~\ref{sec:3.4}). ELBO$_{\text{TDS}}$ mitigates this via its reconstruction term.
\end{itemize}

\begin{figure}
    \centering
    \includegraphics[width=0.95\linewidth]{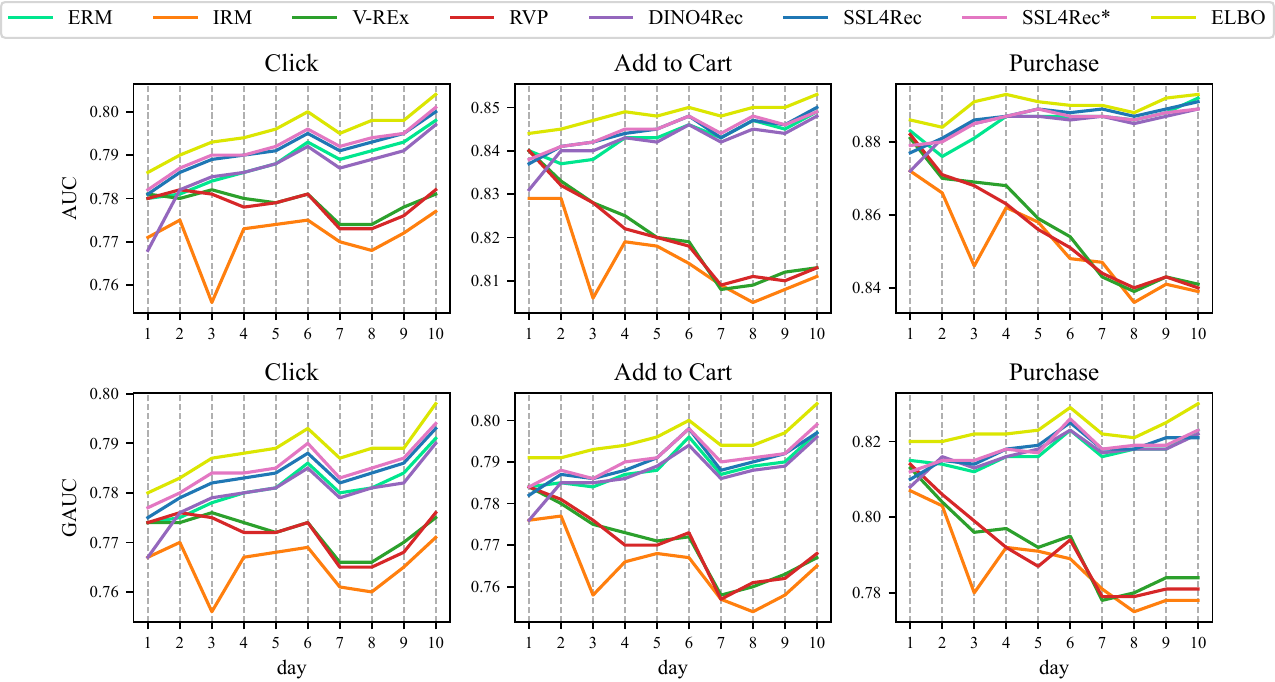}
    \vspace{-2mm}
    \caption{Scaling-over-time capability on Commercial-Small. Each day's checkpoints are tested on next day's data. }
    \label{fig:auc_daily}
\end{figure}

\noindent\textbf{Stability over time.} Figure~\ref{fig:auc_daily} evaluates scaling-over-time under daily incremental training on Commercial-Small. ELBO$_{\text{TDS}}$ maintains a stable uplift over ERM, whereas invariant-learning methods degrade and SSL baselines tend to lose advantage over time.

\subsubsection{\textbf{RQ3: Ablation Study}} We conduct an ablation study to evaluate the individual contributions of different augmentation strategies on Commercial-Small (Table \ref{tab:auc_shopee}). Key insights are as follows: \textbf{(i)}. Each augmentation strategy improves over ERM when combined with ELBO$_{\text{TDS}}$. 
\textbf{(ii)} Statistical feature augmentation has the highest impact, while categorical feature augmentation contributes the least. This suggests that statistical features (e.g., CTRs) fluctuate more significantly over time and thus require stronger regularization. \textbf{(iii)} The combined effect of all augmentations is not strictly additive, meaning the overall performance gain is smaller than the sum of individual contributions. This may be due to the inefficiency in the fusion of augmentations from different feature types, leading to diminishing returns. 
Designing better composition strategies is an interesting direction.

\begin{figure}
\vspace{-2mm}
    \centering
    \includegraphics[width=1.0\linewidth]{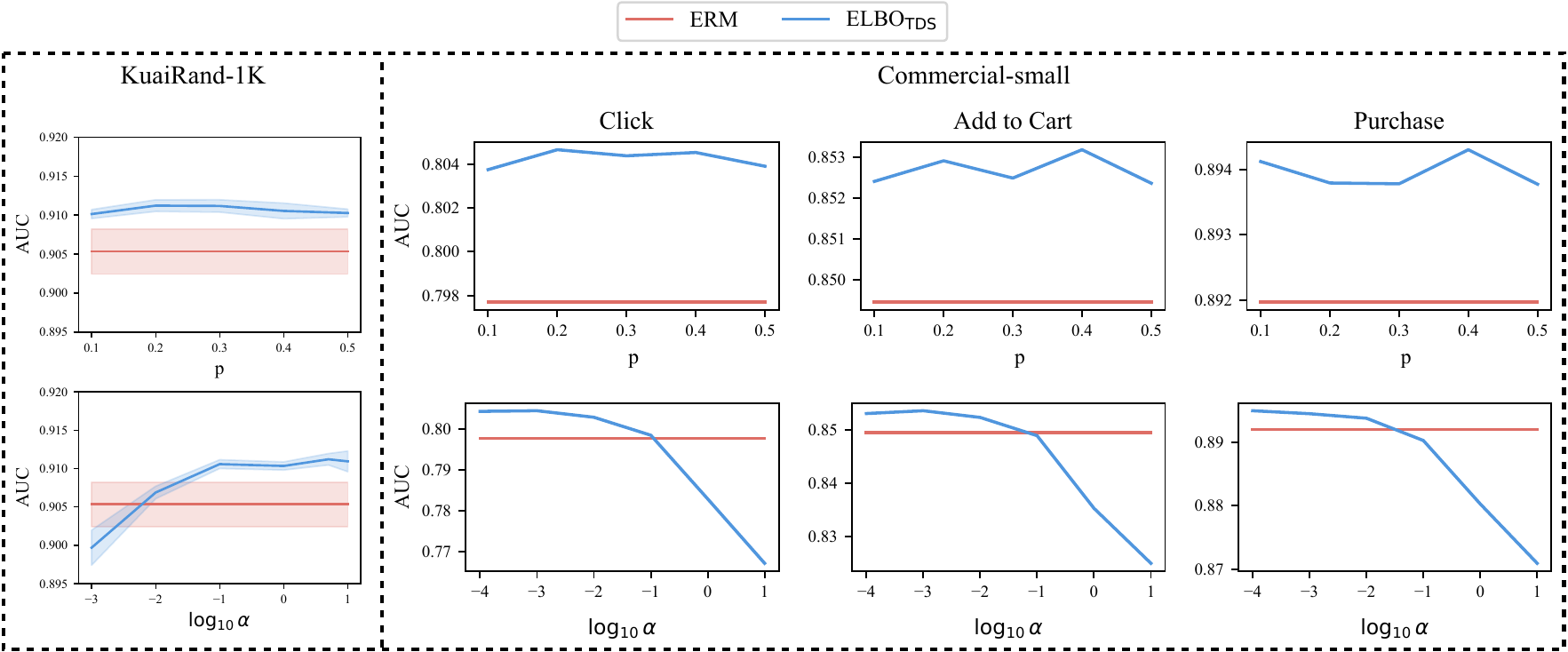}
\vspace{-2mm}
    \caption{Parameter sensitivity of $p$ and $\alpha$ on Kuairand-1K and Commercial-Small.}
    \label{fig:sensitivity-KR}
\end{figure}

\subsubsection{\textbf{RQ4: Hyperparameter Sensitivity}} We analyze the impact of two key hyperparameters: perturbation strength ($p$) and self-supervised loss weight ($\alpha$) across both datasets. The results, shown in Figure \ref{fig:sensitivity-KR}, reveal the following insights: \textbf{(i)} Denser targets (e.g., click) are less sensitive to perturbation strength ($p$), whereas sparser targets (e.g., purchase) are more sensitive. Sparse targets depend on rare positive samples, making them more reliant on effective data augmentation. Insufficient perturbation fails to adequately simulate TDS, while excessive perturbation injects noises, degrading supervision quality. This highlights the importance of tuning $p$ to balance robustness and noise control. \textbf{(ii)} KuaiRand-1K benefits from a higher $\alpha$, while Commercial performs better with a lower $\alpha$. KuaiRand-1K is a small, aggressively sampled dataset, which leads to insufficient fitting for supervised tasks. This also explains why invariant learning does not overfit on KuaiRand-1K. In this case, stronger self-supervision compensates for insufficient labeled data, improving robustness to TDS. Conversely, Commercial-Small contains ample training data, reducing the reliance for self-supervision. A lower $\alpha$ allows the model to focus more on supervised learning, ensuring better adaptation to the latest interactions.

\begin{table}[t]
    \centering
    \caption{Mean daily uplift over ERM on Commercial-Main.}
    \resizebox{\linewidth}{!}{
\begin{tabular}{lcccccc}
\hline
\multirow{2}{*}{\begin{tabular}[c]{@{}l@{}}backbone: MLP\\ loss: CE\&ListNet\cite{ListNet} \end{tabular}} & \multicolumn{2}{c}{Click}           & \multicolumn{2}{c}{Add To Cart}           & \multicolumn{2}{c}{Purchase}         \\
                    & AUC             & GAUC            & AUC             & GAUC            & AUC             & GAUC            \\ \hline
DINO4Rec            & +0.05\%          & +0.01\%          & -0.02\%          & -0.01\%          & -0.02\%          & +0.05\%          \\
SSL4Rec*            & +0.51\%          & +0.34\%          & +0.42\%          & +0.31\%          & +0.34\%          & +0.28\%          \\
ELBO$_{\text{TDS}}$ & \textbf{+0.93\%} & \textbf{+1.02\%} & \textbf{+0.63\%} & \textbf{+0.71\%} & \textbf{+0.69\%} & \textbf{+0.73\%} \\ \hline 

\end{tabular}
}
    
    \label{tab:sp-main}
\end{table}

\subsection{Online Evaluation (RQ5)}
We further evaluate ELBO$_{\text{TDS}}$ in an online setting within a large-scale multi-stage recommendation system. These results serve as supplementary evidence beyond offline next-period generalization.

\noindent{\textbf{Extended training on a large-scale dataset.}}
To better assess long-horizon behavior, we continue training selected methods on a larger dataset until convergence and report the average next-period AUC/GAUC uplift over incremental ERM (Table~\ref{tab:sp-main}). ELBO$_{\text{TDS}}$ maintains a consistent improvement over both SSL baselines, supporting its scalability under prolonged incremental updates.

\noindent\textbf{Online A/B Experiment.}
We run a controlled traffic-split experiment comparing a baseline incremental-training system against the same system equipped with ELBO$_{\text{TDS}}$ at two ranking stages. We report the platform’s primary business-aligned metric (GMV/User) and an independent human-rated relevance metric. Over a two-week window, ELBO$_{\text{TDS}}$ improves GMV/User by \textbf{+2.33\%} and improves human-rated relevance by \textbf{+0.2\%}, indicating consistent gains under real temporal dynamics.

\setlength{\columnsep}{6pt}
\begin{wrapfigure}{r}{0.6\linewidth}
    \vspace{-4mm}
    \centering
    \includegraphics[width=\linewidth]{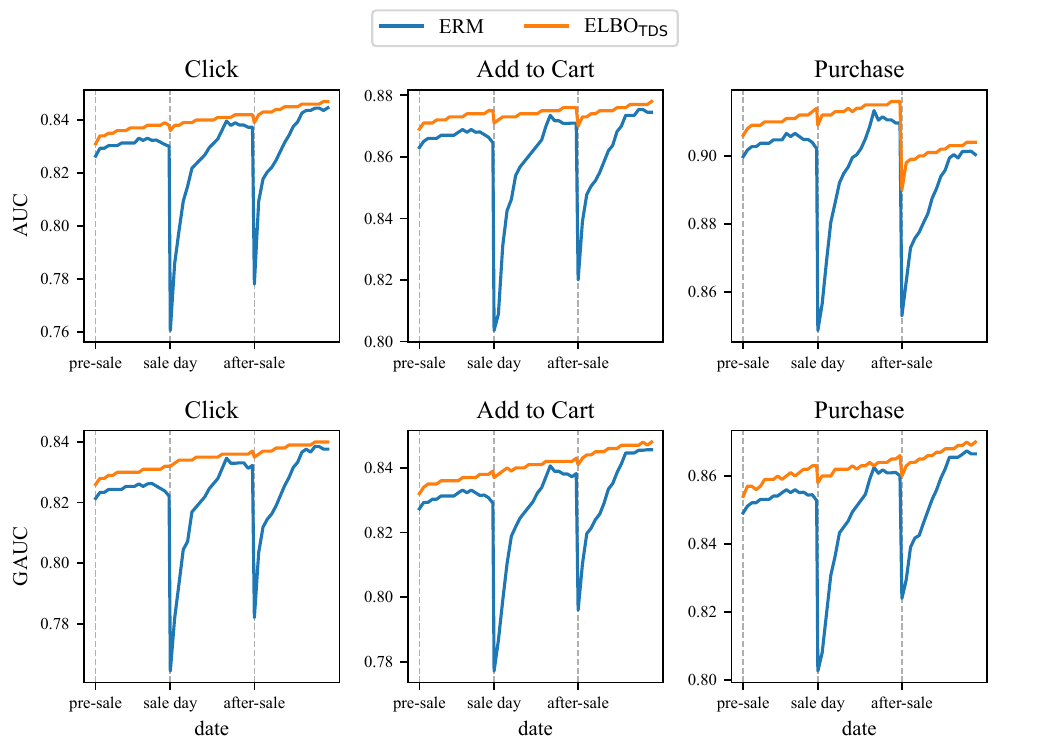}
    \vspace{-6mm}
    \caption{Hourly AUC/GAUC during Commercial's Promotion.}
    \label{fig:promotion-auc}
    \vspace{-4mm}
\end{wrapfigure}
\noindent{\textbf{Robustness under distribution shift.}}
During the online window, a major promotion event induced an abrupt distribution shift. 
Figure~\ref{fig:promotion-auc} shows next-hour AUC/GAUC over the three-day period around the event. Incremental ERM exhibits a sharp degradation at the onset of the shift, while ELBO$_{\text{TDS}}$ remains substantially more stable and recovers faster, suggesting improved robustness without increasing retraining frequency.

\section{Related Work}
Our work is closely related to recommender systems, out-of-distribution (OOD) generalization, and self-supervised learning (SSL).

\noindent\textbf{Industrial Recommender Systems}
typically combine sparse feature embeddings with deep feature interaction modules to predict user behavior\cite{pmlr-v235-zhang24ao}. They rely heavily on engineered categorical, statistical, and sequential features to improve ranking quality \cite{wide&deep,deepfm,MMoE,PLE,airbnb}. Despite strong empirical performance, these systems are usually trained under empirical risk minimization and remain vulnerable to temporal distribution shift.

\noindent\textbf{OOD generalization} 
aims to improve robustness under distribution shift \cite{liu2023outofdistributiongeneralizationsurvey}.
\textit{Unsupervised domain adaptation} \cite{JMLR:v17:15-239,liu2022deep} aligns feature distributions between the source and target domains using unlabeled target data. However, in industrial recommender systems, future distributions are inaccessible. \textit{Invariant learning} seeks representations stable across different environments, enabling generalization to unseen data \cite{arjovsky2020invariantriskminimization,pmlr-v139-krueger21a,yong2021empirical,Sagawa*2020Distributionally}. 
Despite theoretical appeal, they are difficult to apply in our setting because temporal drift is continuous, environment boundaries are ambiguous, and replaying multiple historical windows is expensive in industry-scale pipelines. Our work instead targets next-period temporal generalization without requiring target-time data or historical replay.

\noindent\textbf{Self-supervised learning.}
SSL improves representation learning through augmentation-based objectives \cite{zhang2018mixup,sui2023unleashing,9878654,Rong2020DropEdge:}, including contrastive \cite{9157636,chen2020improvedbaselinesmomentumcontrastive,9711302,10.1007/978-3-031-19809-0_38}, non-contrastive \cite{NEURIPS2020_f3ada80d,9578004,9709990,oquab2023dinov2}, and generative \cite{bizeul2024a,9879206,9880205,gidaris2018unsupervisedrepresentationlearningpredicting} methods.
In recommender systems, SSL has gained significant traction \cite{10.1145/3616855.3635814}. Notably, BUIR \cite{10.1145/3404835.3462935} employs a teacher-student framework akin to BYOL, while SelfCF \cite{10.1145/3591469} adopts a SimSiam-like structure for contrastive learning. SSL4Rec \cite{10.1145/3459637.3481952}, the most relevant to our work, applies InfoNCE regularization to item embeddings to generalize to long-tail items and enhance accuracy.

\section{Conclusion and Discussion}

In this work, we tackle the overlooked problem of temporal distribution shift (TDS) in recommender systems. We introduce ELBO$_{\text{TDS}}$, a probabilistic framework that improves robustness in dynamic, real-world settings. The framework has two core components: (i) a simple yet effective data augmentation scheme that mimics temporal fluctuations to extend the training support; and (ii) an objective ELBO$_{\text{TDS}}$ exploits this extended distribution while avoiding representation collapse. The method is lightweight and plug-and-play, requiring minimal changes to existing recommender architectures and incremental training pipelines. Extensive offline experiments and online A/B tests demonstrate its superiority in practice. 

\noindent\textbf{Limitation.}
Our augmentations are a pragmatic approximation of time-varying factors under TDS, and we do not attempt to explicitly model every potential shift mechanism in a production pipeline. An important direction is to \textit{learn} shift factors directly from data, and to adapt the approximation strategy over time as drift evolves.

\section*{Acknowledgment}
This research is supported by the Lee Kong Chian Fellowship awarded to Yuan Fang by Singapore Management University.



\bibliographystyle{ACM-Reference-Format}
\bibliography{sample-base}

\onecolumn
\appendix
\section{Proofs}

\subsection{Derivation of the ELBO of $\mathop{\max}_{\theta}{\mathbb{E}_{(\vecrv{x}_u,\vecrv{x}_i,\rv{y}_{u,i})\sim p_{\mathcal{T}|\vecrv{s}_u,\vecrv{s}_i,\vecrv{v}_u,\vecrv{v}_i}}[\mathop{\log}p_{\theta}(\vecrv{x}_u,\vecrv{x}_i,\rv{y}_{u,i}|\vecrv{s}_u,\vecrv{s}_i)]}$}
\label{app:proof-elbo}

\begin{proof}
\begin{equation*}
\begin{aligned}
&\mathop{\max}_{\theta}{\mathbb{E}_{(\vecrv{x}_u,\vecrv{x}_i,\rv{y}_{u,i})\sim p_{\mathcal{T}|\vecrv{s}_u,\vecrv{s}_i,\vecrv{v}_u,\vecrv{v}_i}}[\mathop{\log}p_{\theta}(\vecrv{x}_u,\vecrv{x}_i,\rv{y}_{u,i}|\vecrv{s}_u,\vecrv{s}_i)]} \\
=& \mathop{\max}_{\theta, \phi}{\mathbb{E}_{(\vecrv{x}_u,\vecrv{x}_i,\rv{y}_{u,i})\sim p_{\mathcal{T}|\vecrv{s}_u,\vecrv{s}_i,\vecrv{v}_u,\vecrv{v}_i}}[\mathop{\log}p_{\theta}(\vecrv{x}_u,\vecrv{x}_i,\rv{y}_{u,i}|\vecrv{s}_u,\vecrv{s}_i) \int q_{\phi}(\vecrv{z}_u,\vecrv{z}_i|\vecrv{x}_u,\vecrv{x}_i) d\vecrv{z}_u d\vecrv{z}_i]} \\
=&\mathop{\max}_{\theta, \phi}{\mathbb{E}_{(\vecrv{x}_u,\vecrv{x}_i,\rv{y}_{u,i})\sim p_{\mathcal{T}|\vecrv{s}_u,\vecrv{s}_i,\vecrv{v}_u,\vecrv{v}_i}}[\int q_{\phi}(\vecrv{z}_u,\vecrv{z}_i|\vecrv{x}_u,\vecrv{x}_i) \mathop{\log}p_{\theta}(\vecrv{x}_u,\vecrv{x}_i,\rv{y}_{u,i}|\vecrv{s}_u,\vecrv{s}_i) d\vecrv{z}_u d\vecrv{z}_i]} \\
=&\mathop{\max}_{\theta, \phi}{\mathbb{E}_{(\vecrv{x}_u,\vecrv{x}_i,\rv{y}_{u,i})\sim p_{\mathcal{T}|\vecrv{s}_u,\vecrv{s}_i,\vecrv{v}_u,\vecrv{v}_i}}[\mathbb{E}_{\vecrv{z}\sim q_{\phi}(\vecrv{z}_u,\vecrv{z}_i|\vecrv{x}_u,\vecrv{x}_i)}[\mathop{\log}p_{\theta}(\vecrv{x}_u,\vecrv{x}_i,\rv{y}_{u,i}|\vecrv{s}_u,\vecrv{s}_i)]]} \\
=&\mathop{\max}_{\theta, \phi}{\mathbb{E}_{(\vecrv{x}_u,\vecrv{x}_i,\rv{y}_{u,i})\sim p_{\mathcal{T}|\vecrv{s}_u,\vecrv{s}_i,\vecrv{v}_u,\vecrv{v}_i}}[\mathbb{E}_{\vecrv{z}\sim q_{\phi}(\vecrv{z}_u,\vecrv{z}_i|\vecrv{x}_u,\vecrv{x}_i)}[\mathop{\log} \frac{p_{\theta}(\vecrv{x}_u,\vecrv{x}_i,\rv{y}_{u,i}|\vecrv{z}_u,\vecrv{z}_i,\vecrv{s}_u,\vecrv{s}_i)p(\vecrv{z}_u,\vecrv{z}_i|\vecrv{s}_u,\vecrv{s}_i)}{p_{\theta}(\vecrv{z}_u,\vecrv{z}_i|\vecrv{x}_u,\vecrv{x}_i,\rv{y}_{u,i},\vecrv{s}_u,\vecrv{s}_i)} \cdot \frac{q_{\phi}(\vecrv{z}_u,\vecrv{z}_i|\vecrv{x}_u,\vecrv{x}_i)}{q_{\phi}(\vecrv{z}_u,\vecrv{z}_i|\vecrv{x}_u,\vecrv{x}_i)}]]} \\
=&\mathop{\max}_{\theta, \phi}{\mathbb{E}_{(\vecrv{x}_u,\vecrv{x}_i,\rv{y}_{u,i})\sim p_{\mathcal{T}|\vecrv{s}_u,\vecrv{s}_i,\vecrv{v}_u,\vecrv{v}_i}}[\mathbb{E}_{\vecrv{z}\sim q_{\phi}(\vecrv{z}_u,\vecrv{z}_i|\vecrv{x}_u,\vecrv{x}_i)}[\mathop{\log} \frac{p_{\theta}(\vecrv{x}_u|\vecrv{z}_u)p_{\theta}(\vecrv{x}_i|\vecrv{z}_i)p_{\theta}(\rv{y}_{u,i}|\vecrv{z}_u,\vecrv{z}_i)p(\vecrv{z}_u|\vecrv{s}_u)p(\vecrv{z}_i|\vecrv{s}_i)}{p_{\theta}(\vecrv{z}_u,\vecrv{z}_i|\vecrv{x}_u,\vecrv{x}_i,\rv{y}_{u,i},\vecrv{s}_u,\vecrv{s}_i)} \cdot \frac{q_{\phi}(\vecrv{z}_u,\vecrv{z}_i|\vecrv{x}_u,\vecrv{x}_i)}{q_{\phi}(\vecrv{z}_u,\vecrv{z}_i|\vecrv{x}_u,\vecrv{x}_i)}]]} \\
&\text{(based on the data-generating process in Figure~\ref{fig:causal-graph}, }(\vecrv{x}_u,\vecrv{x}_i,\rv{y}_{u,i})\ind (\vecrv{s}_u,\vecrv{s}_i)\,|\,(\vecrv{z}_u,\vecrv{z}_i) , (\vecrv{x}_u,\vecrv{x}_i) \ind \rv{y}_{u,i}\,|\,(\vecrv{z}_u,\vecrv{z}_i), \text{)}\\
=&\mathop{\max}_{\theta, \phi}\mathbb{E}_{(\vecrv{x}_u,\vecrv{x}_i,\rv{y}_{u,i})\sim p_{\mathcal{T}|\vecrv{s}_u,\vecrv{s}_i,\vecrv{v}_u,\vecrv{v}_i}}[\mathbb{E}_{\vecrv{z}\sim q_{\phi}(\vecrv{z}_u,\vecrv{z}_i|\vecrv{x}_u,\vecrv{x}_i)}[\mathop{\log} \frac{p_{\theta}(\vecrv{x}_u|\vecrv{z}_u)p_{\theta}(\vecrv{x}_i|\vecrv{z}_i)p_{\theta}(\rv{y}_{u,i}|\vecrv{z}_u,\vecrv{z}_i)p(\vecrv{z}_u|\vecrv{s}_u)p(\vecrv{z}_i|\vecrv{s}_i)}{q_{\phi}(\vecrv{z}_u,\vecrv{z}_i|\vecrv{x}_u,\vecrv{x}_i)}] \\
&+ D_{\text{KL}}(q_{\phi}(\vecrv{z}_u,\vecrv{z}_i|\vecrv{x}_u,\vecrv{x}_i)\,||\,p_{\theta}(\vecrv{z}_u,\vecrv{z}_i|\vecrv{x}_u,\vecrv{x}_i,\rv{y}_{u,i},\vecrv{s}_u,\vecrv{s}_i))] \\
\geq&\mathop{\max}_{\theta, \phi}{\mathbb{E}_{(\vecrv{x}_u,\vecrv{x}_i,\rv{y}_{u,i})\sim p_{\mathcal{T}|\vecrv{s}_u,\vecrv{s}_i,\vecrv{v}_u,\vecrv{v}_i}}[\mathbb{E}_{\vecrv{z}\sim q_{\phi}(\vecrv{z}_u,\vecrv{z}_i|\vecrv{x}_u,\vecrv{x}_i)}[\mathop{\log} \frac{p_{\theta}(\vecrv{x}_u|\vecrv{z}_u)p_{\theta}(\vecrv{x}_i|\vecrv{z}_i)p_{\theta}(\rv{y}_{u,i}|\vecrv{z}_u,\vecrv{z}_i)p(\vecrv{z}_u|\vecrv{s}_u)p(\vecrv{z}_i|\vecrv{s}_i)}{q_{\phi}(\vecrv{z}_u,\vecrv{z}_i|\vecrv{x}_u,\vecrv{x}_i)}]]} \\
&\text{(with mean-field variational encoder } q_{\phi}(\vecrv{z}_u,\vecrv{z}_i|\vecrv{x}_u,\vecrv{x}_i)=q_{\phi}(\vecrv{z}_u|\vecrv{x}_u)q_{\phi}(\vecrv{z}_i|\vecrv{x}_i) \text{)} \\
=&\mathop{\max}_{\theta, \phi}\mathbb{E}_{(\vecrv{x}_u,\vecrv{x}_i,\rv{y}_{u,i})\sim p_{\mathcal{T}|\vecrv{s}_u,\vecrv{s}_i,\vecrv{v}_u,\vecrv{v}_i}}[\mathbb{E}_{\vecrv{z}_u\sim q_{\phi}(\vecrv{z}_u|\vecrv{x}_u),\vecrv{z}_i\sim q_{\phi}(\vecrv{z}_i|\vecrv{x}_i)}[\mathop{\log} p_{\theta}(\vecrv{x}_u|\vecrv{z}_u) + \mathop{\log}p_{\theta}(\vecrv{x}_i|\vecrv{z}_i) + \mathop{\log} p_{\theta}(\rv{y}_{u,i}|\vecrv{z}_u,\vecrv{z}_i)\\
&-\mathop{\log}q_{\phi}(\vecrv{z}_u|\vecrv{x}_u)-\mathop{\log}q_{\phi}(\vecrv{z}_i|\vecrv{x}_i) + \mathop{\log} p(\vecrv{z}_u|\vecrv{s}_u)+ \mathop{\log} p(\vecrv{z}_i|\vecrv{s}_i)]]\\
\end{aligned}
\end{equation*}
\end{proof}

\subsection{Proof of Proposition~\ref{prop:infonce_upper_bound}}
\label{sec:proof-prop1}
\begin{boxed}
PROPOSITION 1. 
Given the user and item feature $\vecrv{x}_u$ and $\vecrv{x}_i$ with their respective augmented view $\vecrv{X}_u=\{\vecrv{x}_u,\vecrv{x}^{+,0}_u,\vecrv{x}^{-,1}_u,..,\vecrv{x}^{-,k}_u\}$ and $\vecrv{X}_i=\{\vecrv{x}_i,\vecrv{x}^{+,0}_i,\vecrv{x}^{-,1}_i,..,\vecrv{x}^{-,k}_i\}$ the InfoNCE loss upper bounds the following expression:
\begin{equation*}
\mathcal{L}_{\text{InfoNCE}}^{u} + \mathcal{L}_{\text{InfoNCE}}^{i}\geq -\mathbb{E}_{\vecrv{x}_u, \vecrv{x}_i}[\mathbb{E}_{\vecrv{z}_u, \vecrv{z}_i}[\log p(\vecrv{z}_u|\vecrv{s}_u) + \log p(\vecrv{z}_i|\vecrv{s}_i) - \log p(\vecrv{z}_u) - \log p(\vecrv{z}_i)]] + c,
\end{equation*}
where $c=2\log k$.
\end{boxed}

\begin{proof}
We prove the proposition by first proving the upper bound of a single InfoNCE by analogy to \cite{}, then we combine the the two InfoNCE for user and item.

We first derive the single InfoNCE loss as similar with \cite{oord2019representationlearningcontrastivepredictive}. Formally, for a feature $\vecrv{x}$ with their respective augmented view $\vecrv{X}=\{\vecrv{x},\vecrv{x}^{+,0},\vecrv{x}^{-,1},..,\vecrv{x}^{-,k}\}$, where $\{\vecrv{x},\vecrv{x}^{+,0} \}$ denotes the positive pair and $\{\vecrv{x},\vecrv{x}^{-,l} \}_{l=1}^{k}$ denotes the negative pairs. With a mild abuse of notation, we leverage $[\rv{y}=0]$ to indicate the covariate $\vecrv{x}$ being ``positive'', and the InfoNCE is dirived as follows.
\begin{equation*}
\begin{aligned}
&\mathbb{E}_{\vecrv{X}}[\log p(\rv{y}=0|\vecrv{X})]\\
=&\mathbb{E}_{\vecrv{X}}[\log \mathbb{E}_{\vecrv{Z}\sim q(\vecrv{Z}|\vecrv{X})}[ p(\rv{y}=0|\vecrv{Z})]] \\
\ge&\mathbb{E}_{\vecrv{X}}[\mathbb{E}_{\vecrv{Z}\sim q(\vecrv{Z}|\vecrv{X})}[\log p(\rv{y}=0|\vecrv{Z})]] \quad \text{(Jensen's inequality)} \\
=&\mathbb{E}_{\vecrv{X}}[\mathbb{E}_{\vecrv{Z}\sim q(\vecrv{Z}|\vecrv{X})}[\log \frac{p(\vecrv{Z}^{+,-}|\vecrv{z},\rv{y}=0)p(\rv{y}=0)}{\sum_{l=0}^{k}p(\vecrv{Z}^{+,-}|\vecrv{z},\rv{y}=l)p(\rv{y}=l)}]] \quad \text{(Bayes' rule)}\\
=&\mathbb{E}_{\vecrv{X}}[\mathbb{E}_{\vecrv{Z}\sim q(\vecrv{Z}|\vecrv{X})}[\log \frac{p(\vecrv{z}^{+,0}|\vecrv{z})\prod_{r\neq 0} p(\vecrv{z^{r}})}{\sum_{l=0}^{k}p(\vecrv{z}^{l}|\vecrv{z})\prod_{r\neq l} p(\vecrv{z^{r}})}]] \quad \text{($p(\rv{y}=l)=\frac{1}{k+1}, \forall l$ and independence)}\\
=&\mathbb{E}_{\vecrv{X}}[\mathbb{E}_{\vecrv{Z}\sim q(\vecrv{Z}|\vecrv{X})}[\log \frac{p(\vecrv{z}^{+,0}|\vecrv{z})/p(\vecrv{z^{+,0}})}{\sum_{l=0}^{k}p(\vecrv{z}^{l}|\vecrv{z})/ p(\vecrv{z}^{l})}]] \quad \text{(divide by $\prod_{r=0}^{k}p(\vecrv{z^{r}})$)},
\end{aligned}
\end{equation*}

then we end the proof of InfoNCE by parameterizing via similarity functions:
\begin{equation*}
\mathcal{L}_{\text{InfoNCE}}=-\mathbb{E}_{\vecrv{X}}[\mathbb{E}_{\vecrv{Z}\sim q(\vecrv{Z}|\vecrv{X})}[\log \frac{\exp( \text{sim}(\vecrv{z}^{+,0},\vecrv{z}))}{\sum_{l=0}^{k}\exp(\text{sim}(\vecrv{z}^{l},\vecrv{z}))}]].
\end{equation*}

We next prove the upper bound of InfoNCE as similar in \cite{bizeul2024a}:
\begin{equation*}
\begin{aligned}
\mathcal{L}_{\text{InfoNCE}}=&-\mathbb{E}_{\vecrv{X}}[\mathbb{E}_{\vecrv{Z}\sim q(\vecrv{Z}|\vecrv{X})}[\log \frac{p(\vecrv{z}^{+,0},\vecrv{z})/p(\vecrv{z^{+,0}})}{\sum_{l=0}^{k}p(\vecrv{z}^{l},\vecrv{z})/ p(\vecrv{z}^{l})}]] \quad \text{(multiply by $p(\vecrv{z})$)} \\
=&-\mathbb{E}_{\vecrv{X}}[\mathbb{E}_{\vecrv{Z}\sim q(\vecrv{Z}|\vecrv{X})}[\log p(\vecrv{z}|\vecrv{z}^{+,0}) - \log (p(\vecrv{z}|\vecrv{z}^{+,0}) + \sum_{l=1}^{k}p(\vecrv{z}|\vecrv{z}^{-,l}))]] \\
=&\mathbb{E}_{\vecrv{X}}[\mathbb{E}_{\vecrv{Z}\sim q(\vecrv{Z}|\vecrv{X})}[\log (1+\sum_{l=1}^{k}\frac{p(\vecrv{z}|\vecrv{z}^{-,l})}{p(\vecrv{z}|\vecrv{z}^{+,0})})]] \\
\approx&\mathbb{E}_{\vecrv{x}}[\mathbb{E}_{\vecrv{z}\sim q(\vecrv{z}|\vecrv{x})}[\log (1+k\mathbb{E}_{\vecrv{x}^{+,-}}[\mathbb{E}_{\vecrv{z}^{+,-}\sim q(\vecrv{z}^{+,-}|\vecrv{x}^{+,-})}[\frac{p(\vecrv{z}|\vecrv{z}^{-})}{p(\vecrv{z}|\vecrv{z}^{+,0})}]])]] \\
=&\mathbb{E}_{\vecrv{x}}[\mathbb{E}_{\vecrv{z}\sim q(\vecrv{z}|\vecrv{x})}[\log (1+k\frac{p(\vecrv{z})}{p(\vecrv{z}|\vecrv{z}^{+,0})})]] \\
\geq&\mathbb{E}_{\vecrv{x}}[\mathbb{E}_{\vecrv{z}\sim q(\vecrv{z}|\vecrv{x})}[\log k\frac{p(\vecrv{z})}{p(\vecrv{z}|\vecrv{z}^{+,0})}]] \\
=&-\mathbb{E}_{\vecrv{x}}[\mathbb{E}_{\vecrv{z}\sim q(\vecrv{z}|\vecrv{x})}[\log \frac{p(\vecrv{z},\vecrv{z}^{+,0})}{p(\vecrv{z})p(\vecrv{z}^{+,0})}]] + \log k \\
=&-\mathbb{E}_{\vecrv{x}\sim p_{\mathcal{T}|\vecrv{s},\vecrv{v}}}[\mathbb{E}_{\vecrv{z}\sim q(\vecrv{z}|\vecrv{x})}[\log p(\vecrv{z}|\vecrv{s}) - \log p(\vecrv{z})]] + \log k \quad \text{(alter the notation to align with ELBO)}.
\end{aligned}
\end{equation*}

The last step is to combine the InfoNCE for both user $u$ and item $i$:
\begin{equation*}
\begin{aligned}
\mathcal{L}_{\text{InfoNCE}}^{u}+\mathcal{L}_{\text{InfoNCE}}^{i}&\geq -\mathbb{E}_{\vecrv{x}_u}[\mathbb{E}_{\vecrv{z}_u\sim q(\vecrv{z}_u|\vecrv{x}_u)}[\log p(\vecrv{z}_u|\vecrv{s}_u) - \log p(\vecrv{z}_u)]] -\mathbb{E}_{\vecrv{x}_i}[\mathbb{E}_{\vecrv{z}_i\sim q(\vecrv{z}_i|\vecrv{x}_i)}[\log p(\vecrv{z}_i|\vecrv{s}_i) - \log p(\vecrv{z}_i)]] + 2\log k \\
&=-\mathbb{E}_{\vecrv{x}_u, \vecrv{x}_i}[\mathbb{E}_{\vecrv{z}_u, \vecrv{z}_i}[\log p(\vecrv{z}_u|\vecrv{s}_u) + \log p(\vecrv{z}_i|\vecrv{s}_i) - \log p(\vecrv{z}_u) - \log p(\vecrv{z}_i)]] + 2\log k.
\end{aligned}
\end{equation*}
\end{proof}

\subsection{Proof of Eqn.~\eqref{eq:prior_loss}}
\label{app:proof_prior}

\begin{proof}
We omit the user and item subscripts for simplicity. For $J$ augmented
views sharing the same stable factor $\vecrv{s}$, assume that their latent
representations are conditionally independent and follow
\[
p(\{\vecrv{z}^j\}_{j=1}^J \mid \vecrv{s},\vecscalar{\mu})
=
\prod_{j=1}^J
\mathcal{N}(\vecrv{z}^j;\vecscalar{\mu},\sigma^2\mathbf{I}),
\]
where $\sigma$ is fixed and $\vecscalar{\mu}$ is the shared latent center.

The corresponding negative log-likelihood is
\[
-\log p(\{\vecrv{z}^j\}_{j=1}^J \mid \vecrv{s},\vecscalar{\mu})
=
C+\frac{1}{2\sigma^2}
\sum_{j=1}^J
\left\|\vecrv{z}^j-\vecscalar{\mu}\right\|_2^2,
\]
where $C$ is independent of the latent representations.

The maximum-likelihood estimate of the shared center is
\[
\widehat{\vecscalar{\mu}}
=
\argmin_{\vecscalar{\mu}}
\sum_{j=1}^J
\left\|\vecrv{z}^j-\vecscalar{\mu}\right\|_2^2
=
\frac{1}{J}\sum_{j=1}^J\vecrv{z}^j
=
\overline{\vecrv{z}}.
\]

Substituting this estimate gives
\[
-\log p(\{\vecrv{z}^j\}_{j=1}^J
\mid \vecrv{s},\widehat{\vecscalar{\mu}})
=
C+\frac{1}{2\sigma^2}
\sum_{j=1}^J
\left\|\vecrv{z}^j-\overline{\vecrv{z}}\right\|_2^2.
\]

Since $\sigma$ and $J$ are fixed, minimizing this negative log-likelihood
is equivalent, up to a positive constant factor, to minimizing the MSE
between the representations and their mean. Applying this result to the
user and item representations separately yields Eqn.~\eqref{eq:prior_loss}.

\end{proof}

\section{Detailed Experiment Setup}

\subsection{Dataset Processing} 
The statistics of datasets is given in Table~\ref{tab:stat_datasets}. 
For all Commercial and KuaiRand-1K, statistical features are converted into categorical features via bucketization, ensuring consistency across varying feature distributions. Categorical features are processed using embedding lookup tables, 
which transform discrete values into dense vector representations for model training. 
All sequential features are also sequences of categorical features (historical purchased item IDs), and are processed with shared embedding lookup tables.

\subsection{Model Architecture}
\label{app:model_arch}
\noindent\textbf{Offline Experiment.}
We employ a twin-tower architecture, where user and item representations are learned separately and combined via a dot product to generate final predictions. A shared embedding layer is used to convert categorical features into embeddings.
We use a Multi-Layer Perceptron (MLP) as the primary DNN backbone for both KuaiRand-1K and Commercial. 
The detailed model configurations for each dataset are provided in Table~\ref{tab:model_arch}.
For self-supervised learning (SSL) approaches, including SSL4Rec, SSL4Rec*, Dino4Rec, and ELBO$_{\text{TDS}}$, we leverage a linear encoder after the embedding layer to extract latent representations, ensuring effective feature extraction.

\begin{figure*}
    \centering
    \includegraphics[width=0.8\linewidth]{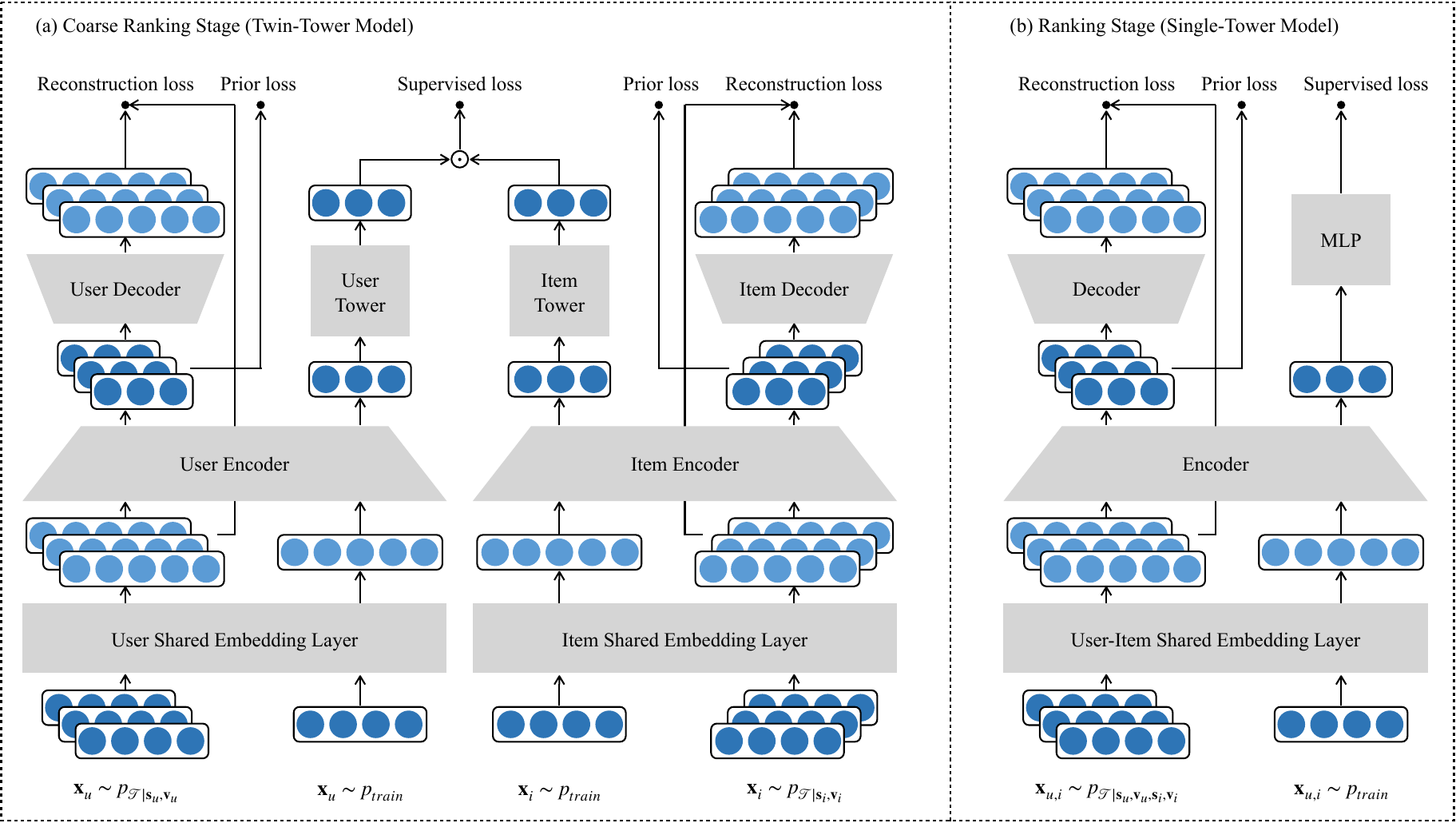}
    \caption{Model architectures on Coarse Ranking Stage and Ranking Stage.}
    \label{fig:two_stage_archs}
\end{figure*}

\noindent\textbf{Online A/B Experiment.} 
We deploy ELBO$_\text{TDS}$ in two key ranking stages of the cascade ranking system: \textit{coarse ranking} and \textit{ranking}.
\textit{Coarse ranking stage} performs a unified ranking on items retrieved from multiple retrieval strategies. The candidate pool is reduced from millions to thousands, and only user-side and item-side features can be used, as item embeddings are precomputed offline and stored in a vector database.
In \textit{ranking stage}, the candidate set is further refined from thousands to hundreds of items. Unlike coarse ranking, this stage allows for the incorporation of user-item cross features to enhance ranking precision. 
The detailed architectures are given in Figure~\ref{fig:two_stage_archs}.


\begin{table*}[]
    \centering
    \caption{Model architectures for Kuairand and Commercial.}
\begin{tabular}{cccccc}
\hline
Dataset                               & methods                                                                          & Backbone & DNN input dim  &DNN                     & \#DNN params  \\ \hline
\multirow{3}{*}{Kuairand single-task} & \begin{tabular}[c]{@{}c@{}}ERM, IRM, \\ V-REx, RVP\end{tabular}                  & MLP      & 1,488 &(128,64,32)             & 210,944   \\ \cline{2-6} 
                                      & \begin{tabular}[c]{@{}c@{}}SSL4Rec, SSL4Rec*, \\ ELBO$_{\text{TDS}}$\end{tabular} & MLP      & 1,488 &Encoder(128)+(128,64,32)    & 243,712   \\ \cline{2-6} 
                                      & \begin{tabular}[c]{@{}c@{}}Dino4Rec, \end{tabular} & MLP      & 1,488 &2*Encoder(128)+(128,64,32)    &434,176   \\ \hline
\multirow{3}{*}{Kuairand multi-task}  & \begin{tabular}[c]{@{}c@{}}ERM, IRM, \\ V-REx, RVP\end{tabular}                  & MLP      & 1,488 &2*(128,64,32)           & 421,888   \\ \cline{2-6} 
                                      & \begin{tabular}[c]{@{}c@{}}SSL4Rec, SSL4Rec*, \\ ELBO$_{\text{TDS}}$\end{tabular} & MLP      & 1,488 &Encoder(128)+2*(128,64,32)  & 296,960   \\ \cline{2-6} 
                                      & \begin{tabular}[c]{@{}c@{}}Dino4Rec, \end{tabular} & MLP      & 1,488 &2*Encoder(128)+2*(128,64,32)    & 487,424   \\ \hline
\multirow{3}{*}{Commercial}               & \begin{tabular}[c]{@{}c@{}}ERM, IRM, \\ V-REx, RVP\end{tabular}                  & MLP  & 3,502 &3*(256,128,16)          & 2,898,432 \\ \cline{2-6} 
                                      & \begin{tabular}[c]{@{}c@{}}SSL4Rec, SSL4Rec*, \\ ELBO$_{\text{TDS}}$\end{tabular} & MLP  & 3,502 &Encoder(512)+3*(256,128,16) & 2,788,352 \\ \cline{2-6} 
                                      & \begin{tabular}[c]{@{}c@{}}Dino4Rec \end{tabular} & MLP  & 3,502 &2*Encoder(512)+3*(256,128,16) & 4,581,376 \\ \hline
\end{tabular}
    \label{tab:model_arch}
\end{table*}

\subsection{Algorithm}
We summarize the implementation of ELBO$_{\text{TDS}}$ in Algorithm~\ref{alg:elbo}.
\begin{algorithm}[t]
\caption{ELBO$_{\text{TDS}}$ Training Algorithm}\label{alg:elbo}
\begin{algorithmic}[1]
\Require train set $\{\vecscalar{x}_u^{(i)},\vecscalar{x}_i^{(i)}, y_{u,i}^{(i)}\}_{i=1}^{N}$, batch size $B$, \#augs $J$.

\For{randomly sampled mini-batch $\{\vecscalar{x}_u^{(i)},\vecscalar{x}_i^{(i)}, y_{u,i}^{(i)}\}_{i=1}^{B}$}

\State $\{\vecscalar{x}_u^{(i,j)},\vecscalar{x}_i^{(i,j)}\}_{j=1}^{J}\gets\mathcal{T}(\vecscalar{x}_u^{(i)},\vecscalar{x}_i^{(i)})$ \quad \zhu{\# augment $J$ views}

\State \zhu{\# self-supervised part}
\For{$j=1,2...,J$}
\State $\vecscalar{\mu}_u^{(i,j)} \gets f_{\phi_{u}}(\vecscalar{x}_u^{(i,j)})$ and  $\vecscalar{\mu}_i^{(i,j)} \gets f_{\phi_{i}}(\vecscalar{x}_i^{(i,j)})$
\State $\vecscalar{z}_u^{(i,j)}\gets\vecscalar{\mu}_u^{(i,j)}+\vecscalar{I}\epsilon_u \quad \epsilon_u\sim\mathcal{N}(0, 1)$
\State $\vecscalar{z}_i^{(i,j)}\gets\vecscalar{\mu}_i^{(i,j)}+\vecscalar{I}\epsilon_i \quad \epsilon_i\sim\mathcal{N}(0, 1)$
\State $\vecscalar{\hat{x}}_u^{(i,j)} \gets f_{\theta_{u}}(\vecscalar{z}_u^{(i,j)})$ and $\vecscalar{\hat{x}}_i^{(i,j)} \gets f_{\theta_{i}}(\vecscalar{z}_i^{(i,j)})$
\EndFor
\State \zhu{\# supervised part}
\State $\vecscalar{\mu}_u^{(i)} \gets f_{\phi_{u}}(\vecscalar{x}_u^{(i)})$ and $\vecscalar{\mu}_i^{(i)} \gets f_{\phi_{i}}(\vecscalar{x}_i^{(i)})$
\State $\hat{y}_{u,i}^{(i)} \gets f_{\theta_{i,j}}(\vecscalar{\mu}_u^{(i)},\vecscalar{\mu}_i^{(i)})$
\State  Minimize in Eqn.~\eqref{eq:overall_loss} $\mathcal{L}$ w.r.t. $(\phi_u,\phi_i,\theta_u,\theta_i,\theta_{u,i})$ 
\EndFor
\State return $(\phi_u,\phi_i,\theta_u,\theta_i,\theta_{u,i})$
\end{algorithmic}
\end{algorithm}



\subsection{Hyperparameter Details}
\label{app:hyper-param}
\begin{table}[]
    \centering
    \caption{Hyper-parameters of ELBO$_{\text{TDS}}$.}
\begin{tabular}{lcccccccc}
\hline
                     & Optimizer & $lr$   & $\alpha$ & $p$ & $r$ & $p_{seq}$ & $p_{cate}$ & $J$ \\ \hline
Kuairand single-task & Adam      & 0.0003 & 5.0      & 0.2 & 2   & 0.2       & 0.2        & 4   \\
Kuairand multi-task  & Adam      & 0.0003 & 10.0     & 0.2 & 2   & 0.2       & 0.2        & 4   \\
Commercial               & Adam      & 0.0005 & 0.001    & 0.2 & 2   & 0.2       & 0.2        & 4   \\ \hline
\end{tabular}
    \label{tab:hyper_params}

    \vspace{-4mm}
\end{table}

For ELBO$_{\text{TDS}}$, we tune the hyper-parameters within the following ranges:
learning rate $lr\in[0.0001, 0.1]$, weight of self-supervised loss $\alpha\in\{0.001, 0.01, 0.1, 1.0, 5.0, 10.0\}$, perturbation strength $p\in\{0.1, 0.2, 0.3, 0.4, 0.5\}$, perturbation magnitude for statistical feature $r\in\{1, 2\}$, dropout rate for sequential feature $p_{seq}\in\{0.1, 0.2, 0.3\}$, embedding dropout rate for categorical feature $p_{cate}\in\{0.1, 0.2, 0.3\}$, number of augmentation views $J\in\{2,3,4\}$, summarized in Table~\ref{tab:hyper_params}.

\subsection{Hardware and Software}
We conduct our experiment for Kuairand on NVIDIA V100 (16 GB GPU) machine with PyTorch, and experiment for Commercial on NVIDIA A30 (24 GB GPU) machine with Tensorflow.

\section{Additional Experimental Results}

\subsection{Result on MMoE backbone}

We conduct extensive experiments using an advanced multi-task learning backbone, MMoE~\cite{MMoE}, on both the Commercial-Small and Kuairand-1K. The AUC and GAUC results are reported in Tables~\ref{tab:mmoe_shopee} and~\ref{tab:mmoe_kuai}. Across both datasets, ELBO$_\text{TDS}$ consistently achieves the highest AUC and GAUC scores, confirming its effectiveness and generalizability across different backbone architectures.

\begin{table}[]
    \centering
    \caption{Multi-task listwise ranking on Commercial-Small with MMoE backbone.}
    \vspace{-2mm}

\begin{tabular}{lcccccc}
\hline
\multirow{2}{*}{\begin{tabular}[c]{@{}l@{}}backbone: MMoE\\ loss: CE\&ListNet\cite{ListNet} \end{tabular}} & \multicolumn{2}{c}{Click}         & \multicolumn{2}{c}{Add-to-Cart}   & \multicolumn{2}{c}{Purchase}      \\
                                & AUC             & GAUC            & AUC             & GAUC            & AUC             & GAUC            \\ \hline
ERM                             & 0.8001          & 0.7919          & 0.8496          & 0.7982          & 0.8895          & 0.8204          \\
IRM                             & 0.7803          & 0.7734          & 0.8083          & 0.7617          & 0.8430          & 0.7844          \\
V-REx                           & 0.7833          & 0.7768          & 0.8160          & 0.7716          & 0.8452          & 0.7893          \\
RVP                             & 0.7832          & 0.7763          & 0.8145          & 0.7680          & 0.8427          & 0.7842          \\
DINO4Rec                        & 0.7973          & 0.7908          & 0.8462          & 0.7940          & 0.8864          & 0.8176          \\
SSL4Rec                         & 0.7959          & 0.7900          & 0.8465          & 0.7940          & 0.8869          & 0.8159          \\
SSL4Rec*                        & 0.7993          & 0.7926          & 0.8475          & 0.7948          & 0.8882          & 0.8194          \\
ELBO$_\text{TDS}$               & \textbf{0.8036} & \textbf{0.7966} & \textbf{0.8538} & \textbf{0.8015} & \textbf{0.8925} & \textbf{0.8251} \\ \hline
\end{tabular}
    \label{tab:mmoe_shopee}
\end{table}

\begin{table}[]
    \centering
    \caption{Single- and multi-task ranking performance on Kuairand-1K with MMoE backbone.}
    \vspace{-2mm}
    \begin{tabular}{lcccccc}
\hline
\multirow{3}{*}{backbone: MMoE} & \multicolumn{2}{c}{single-task}   & \multicolumn{4}{c}{multi-task}                                        \\
                                & \multicolumn{2}{c}{is\_like}      & \multicolumn{2}{c}{is\_like}      & \multicolumn{2}{c}{is\_follow}    \\
                                & AUC             & GAUC            & AUC             & GAUC            & AUC             & GAUC            \\ \hline
ERM                             & 0.9048          & 0.5031          & 0.9035          & 0.4913          & 0.7913          & 0.5022          \\
IRM                             & 0.9036          & 0.5052          & 0.9012          & 0.4924          & 0.7879          & 0.5017          \\
V-REx                           & 0.9055          & 0.5048          & 0.9018          & 0.4925          & 0.7907          & 0.5017          \\
RVP                             & 0.9012          & 0.5046          & 0.8999          & 0.4910          & 0.7771          & 0.5031          \\
Dino4Rec                        & 0.9051          & 0.5002          & 0.9043          & 0.5012          & 0.7922          & 0.4967          \\
SSL4Rec                         & 0.9060          & 0.4946          & 0.9066          & 0.5011          & 0.7934          & 0.4950          \\
SSL4Rec*                        & 0.9061          & 0.4974          & 0.9047          & 0.4937          & 0.7937          & 0.4989          \\
ELBO$_\text{TDS}$               & \textbf{0.9102} & \textbf{0.5092} & \textbf{0.9084} & \textbf{0.5096} & \textbf{0.8241} & \textbf{0.5041} \\ \hline
\end{tabular}
    \label{tab:mmoe_kuai}
\end{table}

\subsection{Sensitivity on \#Augmentations}

\begin{figure}
    \centering
    \includegraphics[width=0.45\linewidth]{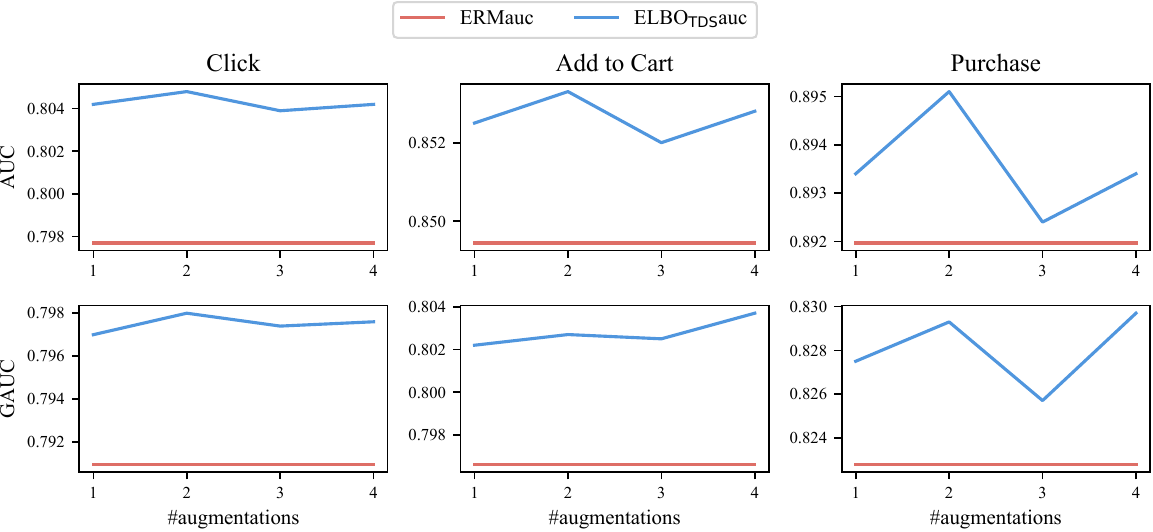}
    \vspace{-2mm}
    \caption{AUC and GAUC w.r.t. \#augmentations on Commercial-Small.}
    \label{fig:auc_augs}
\end{figure}

We further investigate the sensitivity of ELBO$_\text{TDS}$ to the number of augmented views. Figure~\ref{fig:auc_augs} presents the AUC and GAUC results on Commercial-Small with varying numbers of augmentations. We observe that model performance remains stable across 1 to 4 augmentations, indicating ELBO$_\text{TDS}$ is robust to this hyperparameter.


\subsection{Model Efficiency}
\begin{wraptable}{r}{0.4\linewidth} 
\vspace{-4mm} 
\centering
\caption{Train and infer time (ms) per batch on Commercial-Small.}
\vspace{-4mm}
\label{tab:train_infer_time}

\begin{tabular}{lcc}
\hline
                           & Train & Infer \\ \hline
ERM                        & 995        & 204        \\
ELBO$_\text{TDS}$(\#aug=1) & 999        & 190        \\
ELBO$_\text{TDS}$(\#aug=2) & 1122       & 192        \\
ELBO$_\text{TDS}$(\#aug=3) & 1262       & 198        \\
ELBO$_\text{TDS}$(\#aug=4) & 1611       & 207        \\ \hline
\end{tabular}

\vspace{-4mm} 
\end{wraptable}

We further analyze computation cost in terms of FLOPs. Counting one multiply-add as two FLOPs, an $L$-layer MLP with widths $(d_0,\dots,d_L)$ has forward cost $\sum_{l=1}^{L} 2d_{l-1}d_l$. Let $C_{\mathrm{pred}}=\sum_{l=1}^{L_{pred}} 2d_{l-1}d_l$, $C_{\mathrm{enc}}=\sum_{l=1}^{L_{enc}} 2d_{l-1}d_l$, and $C_{\mathrm{dec}}=\sum_{l=1}^{L_{dec}} 2d_{l-1}d_l$ denote the FLOPs of the prediction backbone, encoder, and decoder, respectively. Then ERM requires $\mathrm{FLOPs}^{\mathrm{train}}_{\mathrm{ERM}} \approx \mathrm{FLOPs}^{\mathrm{infer}}_{\mathrm{ERM}} \approx C_{\mathrm{pred}}$, whereas ELBO$_{\text{TDS}}$ requires $\mathrm{FLOPs}^{\mathrm{train}}_{\mathrm{ELBO\text{-}TDS}} \approx C_{\mathrm{pred}} + J(C_{\mathrm{enc}} + C_{\mathrm{dec}})$ and $\mathrm{FLOPs}^{\mathrm{infer}}_{\mathrm{ELBO\text{-}TDS}} \approx C_{\mathrm{enc}} + C_{\mathrm{pred}}$. 
The additional compute of ELBO$_{\text{TDS}}$ is incurred only during training, while the inference FLOPs are independent of $J$. This complexity analysis is consistent with Table~\ref{tab:train_infer_time}.

\end{document}